\documentclass[10pt,cspaper,compsoc]{IEEEtran}
\usepackage{import}
\usepackage{mystyle}
\usepackage{subfiles}
\usepackage{hyperref}
\hypersetup{
	colorlinks   = true,
	urlcolor     = blue,
	linkcolor    = blue%,
}

\usepackage{float}
\usepackage{enumitem}
\usepackage{tabularx}
\usepackage{multirow}
\usepackage{makecell}
\usepackage[switch]{lineno}

\usepackage{ragged2e}
\graphicspath{./figures}

\newcommand{\eg}{e.g.}
\newcommand{\ie}{i.e.}

\newcommand{\G}{\mathcal{G}}

\newcommand{\redtext}[1]{{#1}}

\begin{document}
	% \linenumbers
	\title{Learning Layout and Style Reconfigurable GANs for Controllable Image Synthesis}
	% 	\author{Author Placeholder}
	\author{Wei~Sun and Tianfu~Wu
		\IEEEcompsocitemizethanks{
			\IEEEcompsocthanksitem W. Sun and T. Wu are with the Department of Electrical and Computer Engineering and the Visual Narrative Initiative, North Carolina State University, USA.
			
			E-mail: \{wsun12, tianfu\_wu\}@ncsu.edu
		}
		
	}
	\IEEEtitleabstractindextext{
		\justify
		\begin{abstract}
			With the remarkable recent progress on learning deep generative models, it becomes increasingly interesting to develop models for \textit{controllable} image synthesis from \textit{reconfigurable} \redtext{structured} inputs. This paper focuses on a \redtext{recently} emerged task, \textit{layout-to-image}, \redtext{whose goal is to} learn generative models for synthesizing photo-realistic images from a spatial layout (\ie, object bounding boxes configured in an image lattice) and its style codes (\ie, structural and appearance variations encoded by latent vectors). This paper first proposes an intuitive paradigm for the task, \textit{layout-to-mask-to-image}, \redtext{which learns} to unfold object masks in a weakly-supervised way based on an input layout and object style codes. The layout-to-mask component deeply interacts with layers in the generator network to bridge the gap between an input layout and synthesized images.  Then, this paper presents a method built on Generative Adversarial Networks (GANs) for the proposed layout-to-mask-to-image synthesis with layout and style control at both image and object levels. The controllability is realized by a proposed novel \textit{Instance-Sensitive and Layout-Aware Normalization} (ISLA-Norm) scheme. \redtext{A layout semi-supervised version of the proposed method is further developed without sacrificing performance.} In experiments, the proposed method is tested in the COCO-Stuff dataset and the Visual Genome dataset with state-of-the-art performance obtained.    
		\end{abstract}
		\begin{IEEEkeywords}
			Image Synthesis; Layout-to-Image; Layout-to-Mask-to-Image; Deep Generative Learning; GAN; ISLA-Norm. 
		\end{IEEEkeywords}
	}
	% *** MAKING TITLE ***   
	\maketitle
	
	% \newpage
	% \tableofcontents
	% \newpage
	
	% *** INTRODUCTION ***   
	\IEEEraisesectionheading{\section{Introduction}\label{sec:introduction}}
	\subsection{Motivation and objective}
\begin{figure*} [t]
		\centering
		\includegraphics[width=0.98\textwidth]{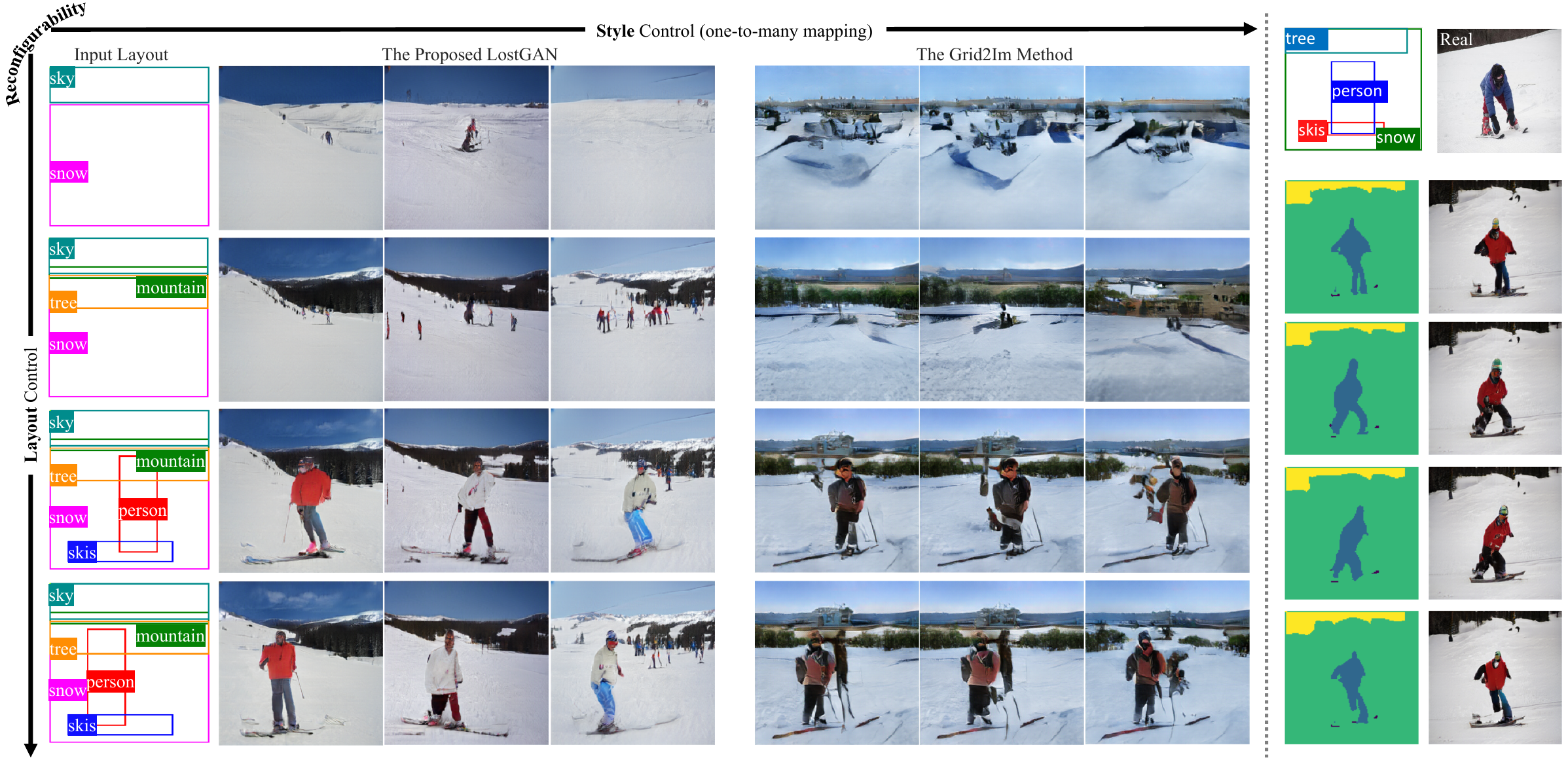}
		\caption{Illustration of controllable image synthesis from reconfigurable spatial layouts and style codes in the COCO-Stuff dataset~\cite{caesar2018coco} at the resolution of $256\times 256$. \underline{\textbf{\textit{On the Left Panel}}}: The proposed method is compared with the prior art, the Grid2Im method~\cite{ashual2019specifying}. \textbf{Each row shows effects of style control}, in which three synthesized images are shown using the same input layout on the left by randomly sampling three style latent codes. \textbf{Each column shows effects of layout control} in terms of consecutively adding new objects (the first three) or perturbing an object bounding box (the last one), while retaining the style codes of existing objects unchanged. \textbf{Advantages of the proposed method:} Compared to the Grid2Im method, (i) the proposed method can generate more diverse images with respect to style control (\eg, the appearance of snow, and the pose and appearance of person). (ii) The proposed method also shows stronger controllability in retaining the style between consecutive spatial layouts. For example, \textit{in the second row}, the snow region is not significantly affected by the newly added mountain and tree regions. Our method can retain the style of snow very similar, while the Grid2Im seems to fail to control. Similarly, \textit{between the last two rows}, our method can produce more structural variations for the person while retaining similar appearance. Models are trained in the COCO-Stuff dataset~\cite{caesar2018coco} and synthesized images are generated at a resolution of $256\times 256$ for both methods.  %The official implementation and pretrained model of Grid2Im are used, which are available at \url{https://github.com/ashual/scene_generation}. 
			\textbf{Note that} the Grid2Im method~\cite{ashual2019specifying} utilizes ground-truth masks in training, while the proposed method is trained without using ground-truth masks, and thus more flexible and applicable in other datasets that do not have mask annotations such as the Visual Genome dataset~\cite{krishna2017visual}. \underline{\textbf{\textit{On the Right Panel:}}} Illustration of the fine-grained control at the object instance level. For an input layout and its real image in the first row, four synthesized masks and images are shown. Compared with the 2nd row,  the remaining three rows show synthesized masks and images by \textit{only} changing the latent code for the Person bounding box. This shows that the proposed method is capable of disentangling object instance generation in an synthesized image at both the layout-to-mask level and  the mask-to-image level, while maintaining a consistent layout in the reconfiguration.  Please see text for details.
		}
		\label{fig:teaser}
	\end{figure*}

	\IEEEPARstart{R}{emarkable} recent progress has been made on both unconditional and conditional image synthesis~\cite{goodfellow2014generative,radford2015unsupervised,zhang2018self,miyato2018spectral,brock2018large,miyato2018cgans,karras2017progressive,karras2018style}. The former aims to generate high-fidelity images from random latent vectors (\eg, sampled from the standard multivariate Gaussian distribution). The latter needs to do so with given conditions satisfied in terms of certain consistency metrics. The conditions may take many forms such as category labels~\cite{odena2017conditional,zhang2018self}, paired or unpaired source images~\cite{kim2017learning,isola2017image,zhu2017unpaired,huang2018multimodal}, semantic maps~\cite{wang2018high,park2019semantic}, text description~\cite{reed2016generative,zhang2017stackgan} and scene graphs~\cite{johnson2018image,ashual2019specifying}. 
	Conditional image synthesis, especially with coarse yet sophisticated and reconfigurable conditions, remains a long-standing problem. As illustrated in Fig.~\ref{fig:teaser}, we shall focus on conditional image synthesis from a spatial layout and its style latent codes, so-called \textit{layout-to-image}~\cite{zhao2020layout2image}. Powerful systems, once developed, can pave a way for computers to truly understand visual patterns and their compositions via a comprehensive and systematic ``analysis-by-synthesis" scheme. Those systems will also enable a wide range of practical applications, \eg, generating high-fidelity data for long-tail scenarios in different vision tasks such as autonomous driving.% and in vision-guided robotic applications. 

	In the layout-to-image synthesis, the layout that a synthesized image needs to satisfy consists of \redtext{a number of} labeled bounding boxes configured in an image lattice (\eg, $256\times 256$ pixels). The style of a synthesized image refers to structural and appearance variations at both image and object levels, which is often encoded by corresponding image and object latent codes. Generating images from a spatial layout represents a sweet spot in conditional image synthesis. Spatial layouts are usually used as intermediate representations for other conditional image synthesis tasks such as text-to-image~\cite{zhang2017stackgan,xu2018attngan} and scene-graph-to-image~\cite{johnson2018image,ashual2019specifying}. And, layouts are more flexible, less constrained and easier to collect than other conditions such as semantic segmentation maps~\cite{isola2017image,wang2018high}. For example, existing object detection benchmarks can be exploited in training. 
	
	The generative learning task of layout-to-image synthesis was recently proposed and only a few work have been proposed in the very recent literature~\cite{johnson2018image,hong2018inferring,zhao2020layout2image,ashual2019specifying}. Although relatively new, it has been well recognized in the computer vision community. For example, the work, Grid2Im by Ashua and Wolf~\cite{ashual2019specifying} won the best paper honorable mentions at ICCV 2019. %~\footnote{\url{ http://iccv2019.thecvf.com/program/main_conference}}. 
	The layout-to-image synthesis task was emerged under the context of remarkable progress made on conditional image synthesis with relatively less complicated conditions such as the class-conditional image synthesis in ImageNet by the BigGAN~\cite{brock2018large}, and the amazing style control for specific objects (e.g., faces and cars) by the StyleGAN~\cite{karras2018style}%~\footnote{We view the StyleGAN as an implicitly conditional image synthesis framework since only one category is usually handled in training}. 
	Despite the big successes achieved by BigGANs and StyleGANs, learning generative models for layout-to-image synthesis entails more research. 
	%For example, we can train a BigGAN to generate cat images, and as long as the generated images look realistic and sharp, we shall think it does a great job. Similarly, we can train a StyleGAN to generate face images, and we shall be happy if realistic and sharp face images are generated with a natural style (\eg, smiling or sad). 
	In addition to realness, generative models for layout-to-image synthesis need to tackle many spatial and semantic  relationships among multiple objects (combinatorial in general). 
	Specifically, learning layout-to-image synthesis requires addressing the problems of learning one-to-many mapping (\ie, one layout covers many plausible realizations in image synthesis to preserve the intrinsic uncertainty), and of handling consistent multi-object generation (\eg, occlusion handling for overlapped bounding boxes and uneven, especially long-tail distributions of objects). Because of those, it is difficult to capture underlying probability distributions defined in the solution space of layout-to-image synthesis.

	This paper is interested in \textit{controllable image synthesis from reconfigurable layouts and style codes}. As illustrated in Fig.~\ref{fig:teaser}, by controllable and reconfigurable, it means a generative model is capable of (i) \textbf{Layout Control} -- the model is adaptive with respect to changes of layouts (\eg, adding new objects), or perturbations of bounding boxes in a given layout, as well as the style codes associated with the changes of spatial layouts, and (ii) \textbf{Style Control} -- the model preserves the intrinsic one-to-many mapping from a given layout to multiple plausible images with sufficiently different structural and appearance styles (\ie, diversity), at both image and object levels (see the right panel of Fig.~\ref{fig:teaser}). Prior arts on  layout-to-image synthesis mainly focus on low resolution ($64\times 64$)~\cite{johnson2018image,zhao2020layout2image}, except for the very recent Grid2Im method~\cite{ashual2019specifying} which can synthesize images at a resolution of $256\times 256$. We further study \textbf{(i) a layout semi-supervised version of the proposed method} without sacrificing the synthesis performance, which use half of the annotated bounding boxes in the training dataset and whose results shed light on some interesting and important directions for developing stronger layout-to-image synthesis methods, and \textbf{(ii) an end-to-end integration of the proposed method with the SPADE in the GauGAN}~\cite{park2019semantic}, which shows the advantage of the proposed ISLA-Norm scheme.

	\subsection{Method overview}
	To learn controllable image synthesis from reconfigurable layouts and style codes, we build on Generative Adversarial Networks (GANs)~\cite{goodfellow2014generative} and present a \textbf{\textit{L}}ay\textbf{\textit{O}}ut- and \textbf{\textit{ST}}yle-based architecture and learning paradigm for GANs. We termed the proposed method \textbf{LostGAN} in our previous conference paper presented at ICCV 2019, entitled \textit{``Image Synthesis from Reconfigurable Layout and Style"}~\cite{LostGAN} . We shall call the conference version LostGAN-V1 and the updated model LostGAN-V2 in this paper. We first give an overview of our LostGAN and then summarize the changes of LostGAN-V2.

\begin{figure} [t]
		\centering
		\includegraphics[width=0.99\linewidth]{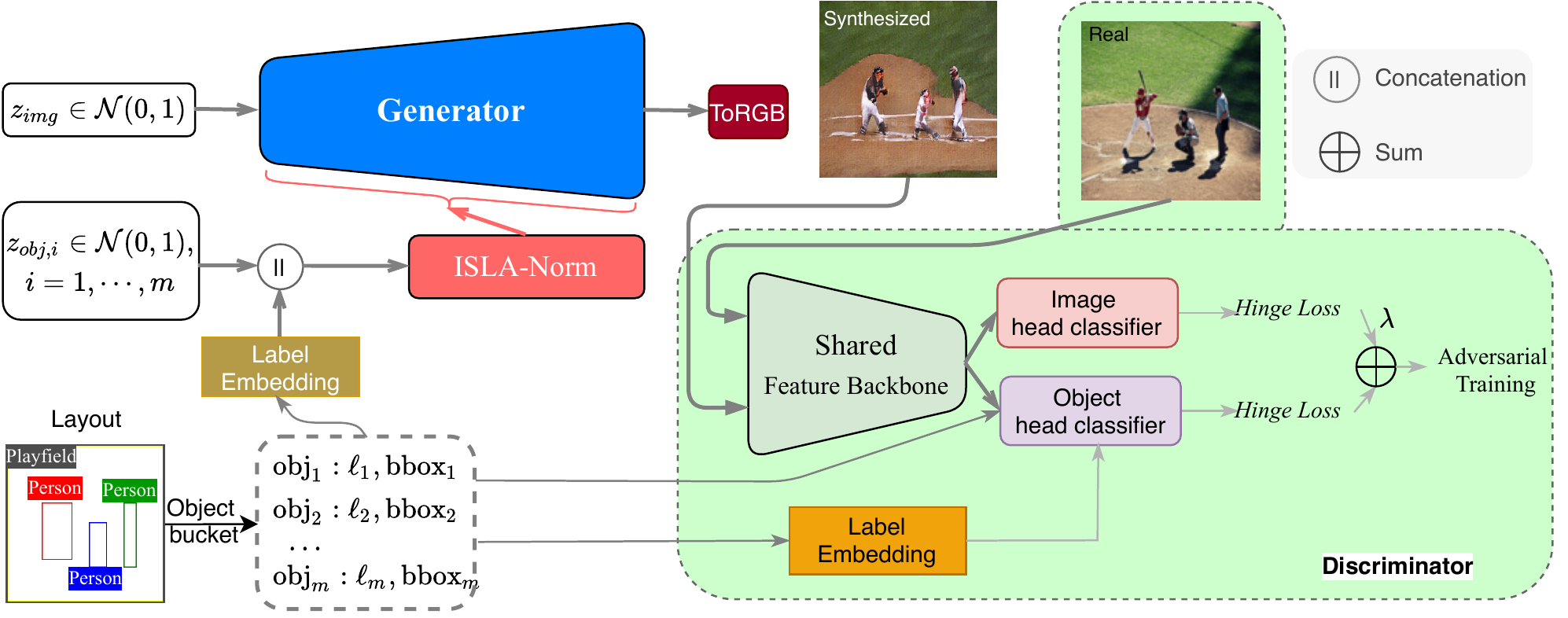}
		\caption{Illustration of the workflow of our proposed LostGANs. Both the generator and discriminator use ResNets~\cite{he2016deep} as backbones. In the generator, ``ToRGB" is a simple module converting the final feature map to RGB images. Our proposed ISLA-Norm and detailed specifications of the generator are explained in Fig.~\ref{fig:norm}. %In the discriminator, the RoIAlign operation~\cite{he2017mask} is used to compute object instance features, which is widely used in two-stage region-based ConvNets for object detection and semantic segmentation. The discriminator is elaborated in Fig.~\ref{fig:D}. %Note that the discriminator  in the green box are not needed after training. See text for details. 
			Best viewed in color.}
		\label{fig:workflow}
	\end{figure}

\begin{figure*} [t]
		\centering
		\includegraphics[width=0.8\linewidth]{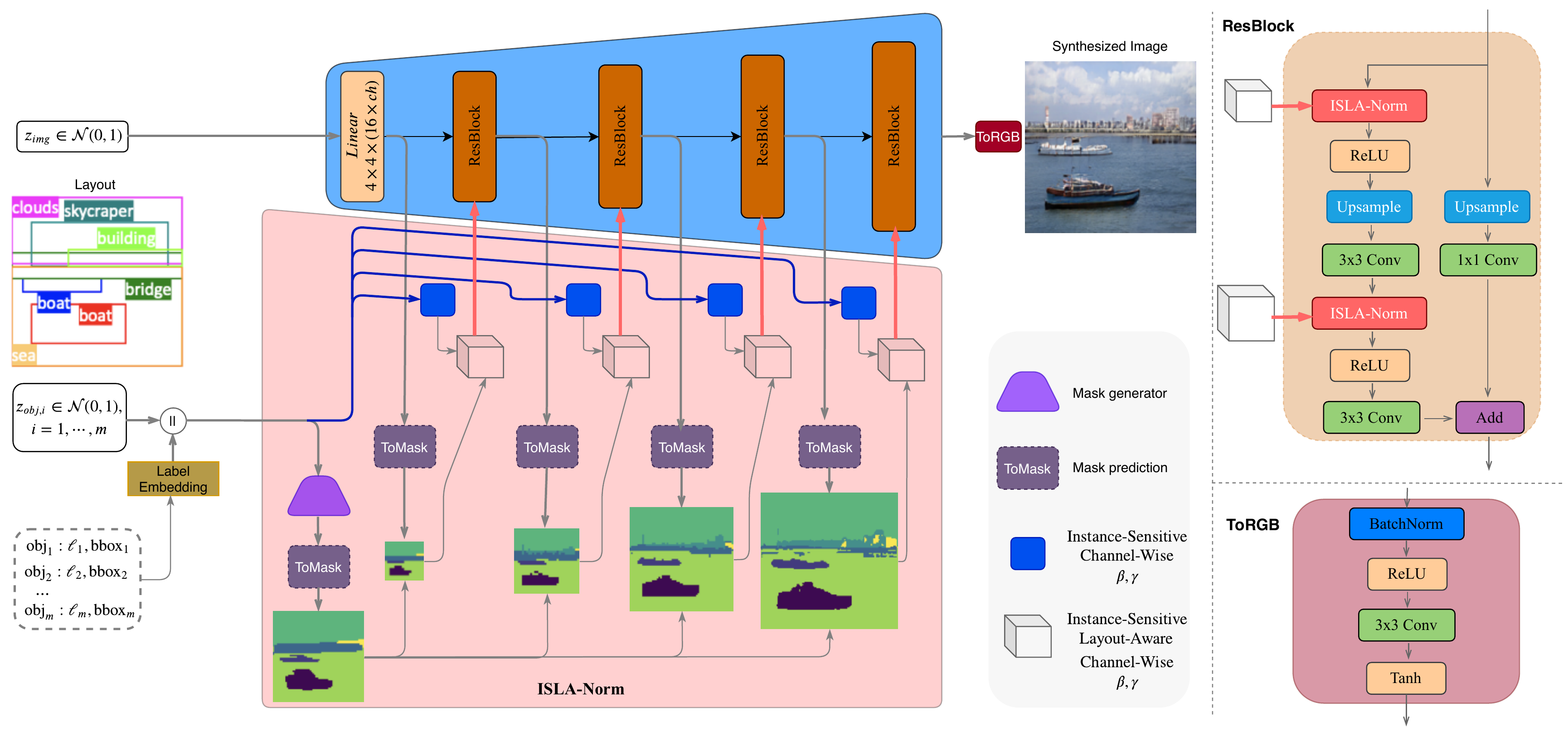}
		\caption{Illustration of our proposed ISLA-Norm for the generator (\textit{left}) and its deployment in a Residual building block (\textit{right-top}). The \textit{right-bottom} illustrates the ``ToRGB" module. The ISLA-Norm realizes the learning of layout-to-mask-to-image synthesis. The masks inferred on-the-fly enrich image synthesis outputs, leading to joint image and label map synthesis. %Note that the masks shown in the figure are computed by a trained model for the given layout. During early stages of the training, the masks are much  noisier. 
			See text for details.}
		\label{fig:norm} 
	\end{figure*}

	\textbf{The proposed LostGAN  addresses the layout-to-image synthesis problem by learning GANs for layout-to-mask-to-image synthesis}. To account for the gap between bounding boxes in a layout and underlying object shapes, learning layout-to-mask is an intuitive and straightforward intermediate step with advantages in two-fold: It induces finer-grained style control of objects in a synthesized image. It also helps decouple learning of object geometry and learning of object appearance. The layout-to-mask generation itself is a relatively easier task than the direct layout-to-image synthesis since object appearance are ignored. In the meanwhile, motivated by the impressive recent progress on conditional image synthesis from semantic label maps~\cite{isola2017image,wang2018high,park2019semantic}, it also makes sense to integrate a layout-to-mask component. If reasonably good object masks can be inferred for an input layout, the learning of mask-to-image synthesis can then leverage the best practice in conditional image synthesis from semantic label maps. A na\"{\i}ve approach is to develop  two-stage generators, which may provide less effective solutions. Instead, we present a single-stage learning paradigm (\ie, using a single generator).  Fig.~\ref{fig:workflow} illustrates the overall workflow of the proposed LostGAN. Fig.~\ref{fig:norm} illustrates the single-stage learning of layout-to-mask-to-image synthesis. 
	
	\textbf{The generator} has three inputs: (i) a spatial layout, $L$ consisting of a number of object bounding boxes in an image lattice, (ii) a latent vector, $z_{img}$ for style control at the image level, and (iii) a concatenation vector between a bucket of object latent vectors, $z_{obj_i}$'s and the label embedding vector of object instances in the layout. The object latent vectors are used for style control of object instances respectively. The generator takes (ii) as its direct input for overall style control, while utilizing a novel feature normalization scheme  for object-level style control based on (iii). 
	
	The object latent vectors are involved in each stage of the generator for better style control and better diversity, similar in spirit to StyleGANs~\cite{karras2018style} (Fig.~\ref{fig:norm}).

	\textbf{The Instance-Sensitive and Layout-Aware Feature Normalization} (ISLA-Norm, Fig.~\ref{fig:norm}) scheme is presented to realize the proposed   layout-to-mask-to-image synthesis pipeline in our LostGANs. As a feature normalization scheme, it consists of two components: feature standardization and feature recalibration. The former is done as the BatchNorm~\cite{ioffe2015batch} in which channel-wise mean and standard deviation are computed in a mini-batch. The latter is different from the BatchNorm. 
	
	Unlike the BatchNorm in which channel-wise affine transformation parameters, $\beta$ (for re-shifting) and $\gamma$ (for re-scaling) are learned as model parameters and shared across spatial dimensions by all instances, our ISLA-Norm first learns \textit{object instance-sensitive channel-wise affine transformations} from the concatenation of object label embedding and object style latent vectors, as shown by the arrows in blue in Fig.~\ref{fig:norm}. This is similar in spirit to the Adaptive Instance Normalization (AdaIN) used in StyleGANs~\cite{karras2018style} and the projection-based conditional BatchNorm used in cGANs~\cite{brock2018large}. Our ISLA-Norm also learns the object masks for objects in an input layout in two pathways: one pathway learns object masks from the concatenation vector between the object label embedding vector and object style latent vectors, which are assembled into a label map, and the other learns a label map from each layer in the generator. A learnable weighted sum of the two label maps are used as the inferred label map at a stage in the generator.  Then, \textit{to obtain fine-grained spatially-distributed multi-object style control for an input layout}, we place the object instance-sensitive channel-wise affine transformations in the learned label map, leading to the instance-sensitive and layout-aware affine transformations for feature recalibration in the generator, as illustrated by the light-grey cubes in Fig.~\ref{fig:norm}. 
	
\begin{figure*} [t]
		\centering
		\includegraphics[width=0.8\linewidth]{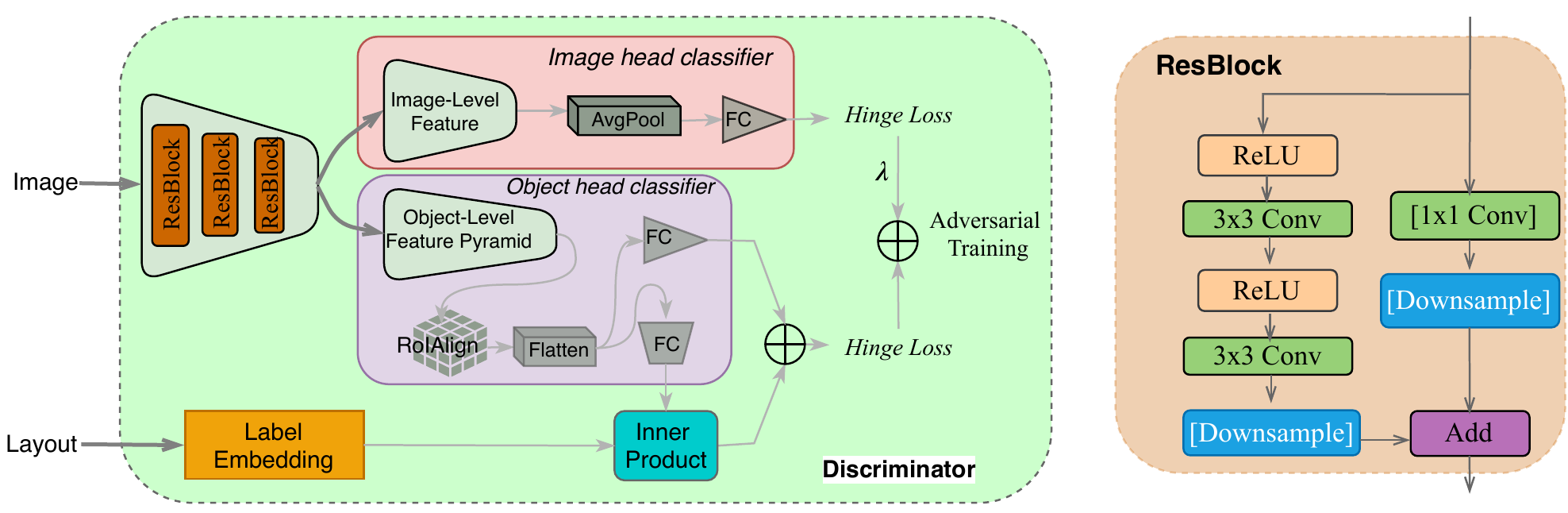}
		\caption{Illustration of the discriminator network (\textit{left}). The shared feature backbone, the image-level feature backbone and the object-level feature pyramid use ResBlocks (\textit{right}). Each of them consists of a number of ResBlocks depending on the target resolution of layout-to-image synthesis (\eg, $256\times 256$). In the ResBlock, ``[op]" means an operation is optional subject to the settings. The object-level feature pyramid is used for placing object instances of different sizes at different feature layers (\eg, smaller bounding boxes placed at lower feature layers as done in the FPN~\cite{FPN}), such that the RoIAlign operation is meaningful.  ``FC" represents a fully-connected layer (with either a scalar output shown by a grey triangle or a vector output shown by a grey trapezoid). ``AvgPool" represents a global channel-wise average pooling over the spatial dimensions.}
		\label{fig:D}
	\end{figure*}

	\textbf{The discriminator} has two inputs: an input image, either fake or real, and the corresponding spatial layout. It consists of three components: (i) a ResNet~\cite{he2016deep} feature backbone, (ii) an image head classifier computing the image realness score
	based on the extracted features (the higher the score is, the more real an image is), and (iii) an object head classifier computing the realness scores for the object instances. The realness score can also be interpreted as playing the role of the negative energy in energy-based models~\cite{lecun2006tutorial,xie2016theory,GrathwohlWJD0S20}, although we do not apply the likelihood-based learning  method in training.
	The feature representation for an object instance is computed by the RoIAlign operator~\cite{he2017mask} using its bounding box in a given layout. Detailed specifications of the discriminator are shown in Fig.~\ref{fig:D}.      
	
	Motivated by the projection-based conditional GANs~\cite{miyato2018cgans} and the practice in  BigGANs~\cite{brock2018large}, a label projection-based score is added to the realness score of each object instance. 
	
	\textbf{The loss function} consists of both image and object adversarial hinge loss terms~\cite{HingeLoss1,HingeLoss2,miyato2018spectral,miyato2018cgans} (balanced by a trade-off parameter, $\lambda$). The hinge loss aims to push the realness score of a synthesized image sufficiently away from that of a real image by a predefined margin. Under the two-player minmax game setting of GANs, the hinge loss works better to enforce both the generator and the discriminator more aggressive in synthesizing images of higher fidelity.

	\textbf{Layout semi-supervised training of LostGANs.} To investigate the possibility of learning from less labels (labeled conditions) in conditional image synthesis and to understand the bottleneck of layout-to-image synthesis, a layout semi-supervised version of LostGAN-v2 is proposed, which obtains comparable performance with the fully-supervised~\footnote{By fully-supervised, it means that each training image has its spatial layout annotated. Accordingly, semi-supervised training means that a portion of the training image does not have spatial layouts annotated (e.g., 50\% training images).} LostGAN-v2. It thus sheds light on some interesting and important directions to develop better layout-to-image synthesis systems. 
	
	\textbf{Summary of Changes.} Compared to our previous  LostGAN-V1~\cite{LostGAN}, the main changes of our LostGAN-V2 are as follows.
	\begin{itemize}[leftmargin=*]
		\itemsep0em
		\item The ISLA-Norm is extended by integrating label maps learned from feature maps at different stages in the generator. Comprehensive experiments are conducted to understand both the effectiveness and benefits of the proposed ISLA-Norm. 
		\item A layout semi-supervised LostGAN-V2 is studied. 
		\item An end-to-end integration of the proposed LostGAN-V2 and the SPADE in the GauGAN~\cite{park2019semantic} is studied.
		\item The experiments are significantly extended by training models at higher resolutions and by comparing with the prior arts including the Grid2Im~\cite{ashual2019specifying} and the GauGAN~\cite{park2019semantic}. 
		\item The paper is thoroughly rewritten with much more details on different aspects of the LostGAN and on the experimental settings, together with new figures of the model. 
		\item Ablation studies are added to analyze the proposed LostGAN and ISLA-Norm. 
	\end{itemize}
	
	\textbf{Our source code and pretrained models of both LostGAN-V1 and V2 have been made publicly available at} \url{https://github.com/iVMCL/LostGANs}.

	\subsection{Related work}
	Generative models have been widely studied in recent years such as Autoregressive models, Variational Autoencoders (VAEs) and Generative Adversarial Networks (GANs). For image generation, Autoregressive models such as PixelRNNs~\cite{oord2016pixel} and PixelCNNs~\cite{van2016conditional} synthesize images pixel by pixel based on conditional distribution over pixels. VAEs~\cite{kingma2013auto,pu2016variational} jointly train an encoder and decoder where the former maps images into latent distribution and the latter generates images based on the latent distribution. GANs~\cite{goodfellow2014generative} are able to synthesize realistic and high resolution images under various settings, including both unconditional~\cite{arjovsky2017wasserstein, karras2017progressive,radford2015unsupervised,karras2018style} and conditional tasks~\cite{odena2017conditional,miyato2018cgans,brock2018large}.  Typically, a GAN consists of a Generator that produces realistic fake images from input (\eg, random noise) and a Discriminator that distinguishes generated images from real ones. More recently, a unified  divergence triangle framework is proposed for joint training of generator model, energy-based model, and inference model for generative tasks~\cite{HanNFHZW19}. Our proposed model is built on GANs and aimed at image synthesis conditioned on coarse semantic layouts.
	
	\textbf{Conditional Image Synthesis.}
	Conditional image synthesis takes additional information (\eg, class information~\cite{mirza2014conditional,odena2017conditional,miyato2018cgans,zhang2018self,brock2018large}, source images~\cite{kim2017learning,zhu2017unpaired,huang2018multimodal,liu2019few}, a text descriptions~\cite{mansimov2015generating,reed2016generative,zhang2017stackgan,li2019object}, scene graphs~\cite{johnson2018image,yikang2019pastegan}) as inputs. How to feed conditional information to a GAN model has been studied in various ways.
	First, in all methods, conditional information are encoded into a vector representation. The encoded condition vector is used differently by different methods. 
	In~\cite{odena2017conditional,reed2016generative}, the encoded condition vector and a sampled latent vector are concatenated as the input to the generator network. 
	In~\cite{reed2016generative,dumoulin2016adversarially,zhang2017stackgan}, the encoded condition vector is utilized by the discriminator by simply concatenating with the input or intermediate feature maps. In~\cite{miyato2018cgans}, projection-based methods exploit conditional information in the discriminator using the inner product between features in the discriminator and the encoded condition vector, which effectively improve the quality of class conditional image generation. 
	In~\cite{de2017modulating,dumoulin2017learned,brock2018large,park2019semantic}, the encoded condition vector is used to control the re-scaling and re-shifting parameters in the BatchNorm~\cite{ioffe2015batch} layers, leading to the conditional BatchNorm.  GauGANs~\cite{park2019semantic} further learn spatially adaptive re-scaling and re-shifting parameters for BatchNorm from an annotated semantic label map. The proposed ISLA-Norm in our previous LostGAN-V1~\cite{LostGAN} is a concurrent work with the feature normalization scheme in GauGANs without resorting to annotated semantic label maps. It learns the layout-to-mask mapping from coarse layout information. The proposed LostGANs also adopt the projection-based methods of exploiting conditional information in the discriminator as done in~\cite{miyato2018cgans}.   
	
	\textbf{Image Synthesis from Layout.}
	Image synthesis from layout has been studied in the very recent literature and proven a difficult task.   The layout-to-image task was first studied in~\cite{zhao2020layout2image} at the resolution of 64$\times$64, which uses a variational autoencoders based network, together with long-short term memory (LSTM), for object feature fusion. In~\cite{yikang2019pastegan}, an external memory bank is introduced, consisting of objects cropped from real images in training, which are retrieved and pasted in generating images from layouts at the resolution of 64$\times$64. In~\cite{johnson2018image,hong2018inferring,hinz2019generating,li2019object}, layout and object information are utilized in text-to-image synthesis or scene-graph-to-image synthesis. \cite{bansal2019shapes} synthesize scene images from given masks by matching context, shape and parts to a stored library. In~\cite{hinz2019generating, hinz2019semantic}, locations of multiple objects are controlled in text-to-image synthesis by adding an extra object pathway in both the generator and discriminator. In~\cite{johnson2018image,hong2018inferring,li2019object}, a two-step approach is used in image synthesis: generating the semantic layout (class label, bounding boxes and semantic mask) from a text description or a scene graph, and synthesizing images conditioned on the predicted semantic layout and text description (if present). However, in~\cite{hong2018inferring,li2019object, ashual2019specifying}, pixel-level instance segmentation annotations are needed in training, while the proposed LostGANs do not require pixel-level annotations and can learn semantic masks in a weakly-supervised way.

	\subsection{Our contributions} 
	This paper makes the following main contributions to the field of conditional image synthesis. 
	\begin{itemize} [leftmargin=*]
		\itemsep0em
		\item It presents a layout- and style-based architecture for GANs (termed LostGANs), which addresses the problem of layout-to-image synthesis by learning layout-to-mask-to-image synthesis. The proposed LostGANs realize controllable image synthesis from reconfigurable layouts and styles. The proposed LostGANs can be trained in a layout fully-supervised way or a layout semi-supervised way. The outputs of LostGANs include both image and semantic label map synthesis. %Our proposed LostGANs integrate the best practices in state-of-the-art conditional and unconditional GANs and two-stage R-CNN based object detection and semantic segmentation systems such as Mask-RCNN. 
		\item It presents an object instance-sensitive and layout-aware feature normalization scheme (termed ISLA-Norm), which explicitly accounts for the joint learning of layout-to-mask generation and spatially-distributed feature recalibration at an object mask level. The ISLA-Norm shows better performance than the SPADE scheme in the GauGAN~\cite{park2019semantic} in the layout to image synthesis task.
		\item It can synthesize images at a resolution of up to $512\times 512$ and shows state-of-the-art performance in terms of the Inception Score~\cite{salimans2016improved}, the Fr\`echet Inception Distance~\cite{FID}, the Diversity Score base on the LPIPS metric~\cite{zhang2018unreasonable}, the classification accuracy score~\cite{ravuri2019classification} and Faster-RCNN~\cite{FasterRCNN} based object detection Average Precision (AP) on two widely used datasets, the COCO-Stuff~\cite{caesar2018coco} and the Visual Genome~\cite{krishna2017visual}. 
		% The diversity score is significantly improved by $133\%$ and $100\%$ relative increase respectively. 
	\end{itemize}
	
	\subsection{Paper organization}
	In the remainder of this paper, Section~\ref{sec:method} presents the problem formulation of layout-to-image and technical details of our proposed LostGANs and ISLA-Norm. Section~\ref{sec:experiments} shows the experimental settings, quantitative and qualitative results, together with ablation studies. Section~\ref{sec:conclusion} concludes this paper with discussions on some directions for future work. 
	
	\section{The Proposed Method} \label{sec:method}
	% In this section, we first define the problem of layout-to-image synthesis. Then, we present details of our LostGANs and ISLA-Norm. 
	
	\subsection{Problem formulation}
	Denote by $\Lambda$ an image lattice (\eg, $256\times 256$) and by $I$ an image defined on the lattice. Let $L=\{(\ell_i, bbox_i)_{i=1}^m\}$ be a layout consisting of $m$ labeled bounding boxes, where a label $\ell_i\in \mathcal{C}$ (\eg, $|\mathcal{C}|=171$ in the COCO-Stuff dataset~\cite{caesar2018coco}), and a bounding box $bbox_i\subseteq \Lambda$. Different bounding boxes may overlap and thus have undetermined partial-order of occlusions. 
	
	Let $z_{img}$ be the latent code controlling the image style and $z_{obj_i}$ the latent code controlling the object instance style for $(\ell_i, bbox_i)$. The latent codes are often randomly sampled from the standard multivariate Gaussian distribution, $\mathcal{N}(0,1)$ under the i.i.d. setting. Denote by $Z_{obj}=\{z_{obj_i}\}_{i=1}^m$ the set of object instance style latent codes.
	Image synthesis from layout and style is to learn a mapping from a given input $(L, z_{img}, Z_{obj})$ to a synthesized image $I^{syn}$,
	\begin{equation}
	I^{syn} = \G(L, z_{img}, Z_{obj}; \Theta_{\G}), \label{eq:generator}
	\end{equation}
	where $\Theta_\G$ represents the parameters of the generation function. In general, a generator network $\G(\cdot)$ is expected to capture the underlying conditional data distribution $p(I|L, z_{img}, Z_{obj};\Theta_\G)$ in a high-dimensional space. While straightforward for synthesizing images (using a single phase of forward computation), the generator $\G(\cdot)$ involves a challenging inference step entailed in estimating the model parameters, that is to compute the latent codes for a real image $I^{real}$ by sampling the posterior distribution, $p(z_{img}, z_{obj_1},\cdots, z_{obj_m}|I^{real}, L)$. To mitigate the difficulty of the posterior inference, GANs propose an adversarial training paradigm which exploits an extra discriminator~\cite{goodfellow2014generative} under a two-player minmax game setting.
	
	\textbf{Reconfigurability of a generator network $\G(\cdot)$.} In this paper, we are interested in three aspects as follows:
	\begin{itemize}[leftmargin=*]
		\itemsep0em
		\item \textit{Image style reconfiguration:} For a fixed layout $L$, is the generator $\G(\cdot)$  capable of synthesizing images with different styles for different $(z_{img}, Z_{obj})$ samples, while retaining the layout configuration conditioned on the given $L$?
		\item \textit{Object style reconfiguration:} For a fixed input tuple  $(L, z_{img}, Z_{obj})$ except for one object style latent code $z_{obj_i}\in Z_{obj}$,  is the generator $\G(\cdot)$ capable of generating consistent images with different styles for the object  $(\ell_i, bbox_i)$ using different $z_{obj_i}$ samples, while retaining the object configuration conditioned on the given $L$ and the styles of the remaining objects?
		\item \textit{Layout reconfiguration:} Given an input tuple $(L, z_{img}, Z_{obj})$, is the generator $\G(\cdot)$ capable of generating consistent images for different $(L^+, z_{img}, Z_{obj}^+)$'s, where $L^+$ has a newly added object instance or just changes the location and/or the label of an existing bounding box? When a new object is added, a new $z_{obj}$ is added in $Z_{obj}^+$. When only the bounding box location changes, all latent codes are kept unchanged (\ie, $Z_{obj}^+=Z_{obj}$).  
	\end{itemize}
	It is a challenging problem to address the three aspects by learning a single generator network $\G(\cdot)$. Intuitively, it might be difficult for well-trained artistic people to do so at scale (\eg, the 171 categories in the COCO-Stuff dataset). 
	
	\subsection{The LostGAN}
	\subsubsection{The generator network}
	As illustrated in Fig.~\ref{fig:norm}, the generator $\G(\cdot)$ consists of a linear full-connected (FC) layer, followed by a number of residual building blocks (ResBlocks)~\cite{he2016deep} depending on the target resolution of image synthesis, and a ``ToRGB"  module  outputting a synthesized image. Detailed network architectures for different image synthesis resolutions are referred to our Github repository.

	\subsubsection{The ISLA-Norm}\label{sec:isla-norm}
	There are two ISLA-Norm modules in a ResBlock (the right-top of Fig.~\ref{fig:norm}).
	Denote by $\mathbf{x}$ an input 4D feature map of ISLA-Norm, and $x_{n,c,h,w}$ the feature response at a position $(n, c, h, w)$ (using the convention order of axes for batch, channel, and spatial height and width). We have $n\in [0, N-1], \, c\in [0, C-1],\, h\in [0, H-1],\, w\in [0, W-1]$, where $N$ is the mini-batch size or the accumulated size of synchronized mini-batches, and $C, H, W$ depend on the stage of a ResBlock. 
	
	\textbf{Feature Standardization.} 
	Our ISLA-Norm first computes the channel-wise mean and standard deviation as done in the BatchNorm~\cite{ioffe2015batch}. 
	In training, ISLA-Norm first normalizes $x_{n,c,h,w}$ by,
	\begin{equation}
	\hat{x}_{n,c,h,w} = \frac{x_{n,c,h,w} - \mu_c}{\sigma_c},
	\end{equation}
	where the channel-wise batch mean $\mu_c = \frac{1}{N\cdot H\cdot W} \sum_{n,h,w} x_{n,c,h,w}$ and  standard deviation $\sigma_c=\sqrt{\frac{1}{N\cdot H\cdot W} \sum_{n,h,w} (x_{n,c,h,w}-\mu_c)^2 +\epsilon}$ ($\epsilon$ is a small positive constant for numeric stability). 
	
	\textbf{Feature Recalibration.} 
	In the BatchNorm~\cite{ioffe2015batch}, the recalibration is done by learning channel-wise affine transformations, consisting of the re-scaling paramter, $\gamma_c$'s and the re-shifting parameters, $\beta_c$'s. We have,
	\begin{equation}
	\Tilde{x}_{n,c,h,w}^{BN} = \gamma_{c}\cdot \hat{x}_{n,c,h,w} + \beta_{c}.
	\end{equation}

	Our ISLA-Norm learns instance-sensitive and layout-aware affine transformation parameters, $\gamma_{n,c,h,w}$'s and $\beta_{n,c,h,w}$'s, and we have, 
	\begin{equation}
	\Tilde{x}_{n,c,h,w} = \gamma_{n,c,h,w}\cdot \hat{x}_{n,c,h,w} + \beta_{n,c,h,w},
	\end{equation}
	where both $\gamma_{n,c,h,w}$'s and $\beta_{n,c,h,w}$ are functions of $(L, z_{img}, Z_{obj})$. Thus, \textit{the resulting recalibrated features $\Tilde{x}_{n,c,h,w}$'s are sensitive to both the layout and the image and object style codes, which leads to layout and style reconfigurable image synthesis}.  
	
	\textbf{Computing $\gamma_{n,c,h,w}$ and $\beta_{n,c,h,w}$}. Without loss of generality, we show how to compute the gamma and beta parameters for one sample, i.e., $\gamma_{c,h,w}$ and $\beta_{c,h,w}$. 
	As shown in the left of Fig.~\ref{fig:norm}, we have the following six components. 
	
	\textit{i) Label Embedding.} We use one-hot label vector for the $m$ object instances in a layout $L$, which results in a one-hot label matrix, denoted by $Y$, of the size $m\times d_{\ell}$, where $d_{\ell}$ is the number of object categories (\eg, $d_{\ell}=171$ in the COCO-Stuff dataset). Label embedding is to learn a $d_{\ell}\times d_e$ embedding matrix, denoted by $W$, to compute the vectorized representation for labels, 
	\begin{equation}
	\mathbb{Y} = Y \cdot W, \label{eq:embedding}
	\end{equation}
	where $\mathbb{Y}$ is a $m\times d_e$ matrix and $d_e$ represents the embedding dimension (\eg, $d_e=128$ in our experiments). %We use the built-in implementation such as the {\tt nn.Embedding} module in PyTorch.
	
	\textit{ii) Joint Label and Style Encoding.}
	We sample from the standard Gaussian distribution the object style latent codes $Z_{obj}$ which is a $m\times d_{obj}$ noise matrix (\eg, $d_{obj}=128$ in our experiments). Let $\mathbb{S}$ be the joint label and style encoding, 
	\begin{equation}
	\mathbb{S} = (\mathbb{Y}, Z_{obj}), \label{eq:label_style_encoding}
	\end{equation}
	which is a $m\times (d_e+d_{obj})$ matrix. So, the object instance style depends on both the label embedding (semantics) and i.i.d. latent codes (accounting for style variations).

	\textit{iii) Mask Generation from $\mathbb{S}$}. We first generate an mask for each object instance in a layout $L$ at a predefined size, $s\times s$ (\eg, $s=32$) individually. Then, we resize the generated masks to the sizes of corresponding bounding boxes at a ResBlock stage in the generator. 
	\begin{itemize}[leftmargin=*]
		\itemsep0em
		\item The mask generation process consists of two components: one is a simplified generator model (the small trapezoid in purple in Fig.~\ref{fig:norm}), and the other is a simple ``ToMask" operation with the output tensor of the size $m\times s\times s$, representing a $s\times s$ mask for each of the $m$ object instances. This canonical size of object masks enable our model to handle aspect ratio changes of bounding boxes in image synthesis. Their detailed specifications are referred our Github repository. %The former composes a linear FC layer projecting $\mathbb{S}$ to a tensor with a resolution of $4\times 4$ (after reshaping), and a few stages of Conv3x3+BatchNorm+ReLU (where Conv3x3 refers to a convolution operation with a kernel of spatial dimensions, $3\times 3$). The ``ToMask" operation is then implemented by Conv3x3+Sigmoid, whose output is $m\times s\times s$, representing a $s\times s$ mask for each of the $m$ object instances. Based on the mask size $s$, we also upsample the feature map after a Conv3x3+BatchNorm+ReLU by a factor of $2$ for a number of stages as needed. 
		\item After resizing the generated $m$ object instance mask and placing them back into the layout at the spatial resolution $(H, W)$ of a ResBlock stage, we obtain a mask tensor of dimensions $(m, H, W)$, denoted by $\mathbb{M}_{\mathbb{S}}$, each slice of which has zeros outside the corresponding bounding box, $bbox_i$.  For the visualization purpose (\eg, those shown in Fig.~\ref{fig:norm}), we use $\arg\max$ across the $m$ channels of $\mathbb{M}_{\mathbb{S}}$ to assign the label index for a pixel occupied by more than one objects due to occlusions. 
	\end{itemize}
	
	\textit{iv) Mask updating using the feature maps in a generator}. For a ResBlock stage, we learn a mask from its input feature map using a simple ``ToMask" operation implemented by Conv3x3+Sigmoid, where the out channel of the Conv3x3 kernel is $d_{\ell}$ (\ie, the number of categories in a dataset). The mask is represented by a tensor of sizes $(d_{\ell}, H, W)$. We clip the mask based on the layout by keeping values unchanged within the bounding boxes of the object instances in a layout and zeroing out the remainder. Denote by $\mathbb{M}_{\mathbb{F}}(L)$ the mask tensor of sizes $(m, H, W)$ after the clipping (omitting the index for a ResBlock in the generator). For the second ISLA-Norm module in a ResBlock (the right-top of Fig.~\ref{fig:norm}), we upsample $\mathbb{M}_{\mathbb{F}}(L)$ by a factor of $2$.

	\textit{v) Object instance-sensitive channel-wise affine transformation parameters.} They are learned from the joint label and style encoding $\mathbb{S}$. We adopt a linear projection with a learnable $(d_e+d_{obj})\times 2C$ projection matrix $\mathcal{A}$, where $C$ is the number of channels, and we have, 
	\begin{equation}
	\mathcal{T} = \mathbb{S}\cdot \mathcal{A}, \label{eq:jointstyle}
	\end{equation}
	which is a matrix of sizes $(m, 2C)$. Let $\mathcal{T}_{\beta}$ and $\mathcal{T}_{\gamma}$ be the column-wise first and second half of $\mathcal{T}$. We unsqueeze both $\mathcal{T}_{\beta}$ and $\mathcal{T}_{\gamma}$ to the size of $(m, C, H, W)$ by replicating values across the spatial dimensions. Learning the affine transformation parameters in this way leads to stronger style control and better diversity of our LostGAN than other layout-to-image methods, since the style latent codes get involved in every stage of the generator, rather than being used as input only to the first stage of the generator in other  layout-to-image methods. 
	
	\textit{vi) Computing the ISLA $\gamma_{c,h,w}$ and $\beta_{c,h,w}$.} We first unsqueeze the two masks, $\mathbb{M}_{\mathbb{S}}$ and $\mathbb{M}_{\mathbb{F}}(L)$, to the sizes $(m, C, H, W)$ by replicating $C$ channels. Then, we have, 
	\begin{align}
		\gamma_{c,h,w} = \frac{1}{M_{c, h,w}}\sum_{i=1}^m \mathbb{M}(i, c, h, w) \times \mathcal{T}_{\gamma}(i, c, h, w), \label{eq:isla-gamma}\\
		\beta_{c,h,w} = \frac{1}{M_{c, h,w}}\sum_{i=1}^m \mathbb{M}(i, c, h, w) \times \mathcal{T}_{\beta}(i, c, h, w),  \label{eq:isla-beta}
	\end{align}
	where $\mathbb{M}(\cdot) = [(1-\alpha) \cdot\mathbb{M}_{\mathbb{S}}+ \alpha \cdot  \mathbb{M}_{\mathbb{F}}(L)](\cdot)$ with $\alpha$ being a learnable weight to balance the two masks, and $M_{c,h,w}=\sum_{i=1}^m \mathbb{M}(i, c, h, w)$ if  the pixel $(h,w)$ is occupied by multiple object bounding boxes, otherwise $M_{c, h,w}=1$. 
	%We note that unlike the $\arg\max$ operation that we used for the overlapping regions in visualizing the learned masks, the average is used in computing the spatially-distributed affine transformations. 
	
	\textbf{Handling Background.} To account for the situation in which all object instances do not occupy the entire image lattice (e.g., in the VG dataset~\cite{krishna2017visual}), we introduce a background class $\ell_0$ with $bbox_0=\Lambda$. %The first five steps can be used in a straightforward way. In the last step, we use the background $\gamma$'s and $\beta$'s only for regions which are not covered by any given object instances. Background $\gamma$'s and $\beta$'s are not used in the average for foreground regions.   

	\textbf{Why does the ISLA-Norm help image synthesis?} In sum, one the one hand, as shown by the blue rounded squares in Fig.~\ref{fig:norm} and Eqn.~\ref{eq:label_style_encoding} and Eqn.~\ref{eq:jointstyle}, each layer in the generator network directly learns object instance sensitive channel-wise feature recalibration parameters. So, the object style latent codes $Z_{obj}$ have direct impacts at each layer, unlike the image style latent code $z_{img}$ which may have degenerated impacts on the later layers in the generator in the sense that different $z_{img}$'s may result in very similar images (\ie, less powerful style control). 
	
	On the other hand, the feature recalibration parameters $\gamma_{n,c,h,w}$ and $\beta_{n,c,h,w}$ are further modulated by the learned semantic label map. The semantic label map accounts for two information pathways: one is from the joint embedding of object labels and object style codes (Eqn.~\ref{eq:label_style_encoding}), and the other from the features in the previous layer (see the `ToMask' module in Fig.~\ref{fig:norm}). The second pathway is to learn the residual label map w.r.t. the ouput from the first pathway. These enable the feature recalibration parameters $\gamma_{n,c,h,w}$ and $\beta_{n,c,h,w}$ to deeply interact with object style codes and the intermediate features (computed from the image latent code) in the generator network. 
	
	By doing those, our ISLA-Norm shares the spirit with the AdaIN in StyleGANs~\cite{karras2018style}, but realizes strong style control in a fine-grained spatially-adaptive way. And, our ISLA-Norm further distinguishes itself from the concurrent work of GauGANs~\cite{park2019semantic} which only use $z_{img}$ in controlling styles and resort to ground-truth masks for fine-grained feature recalibration. For an in-depth comparison with GauGANs, we further investigate the integration between the proposed LostGANs and the GauGANs.  
	
	\subsubsection{Comparing with the SPADE module of GauGANs}\label{sec:compare_spade}
	% The GauGAN~\cite{park2019semantic} is one of the state-of-the-art semantic-map-to-image synthesis methods. The GauGAN and our LostGAN-V1~\cite{LostGAN} are concurrent works on developing spatially-adaptive feature normalization schema for condition controllability in conditional GANs. The GauGAN utilizes annotated semantic label maps in training and inference, while our LostGANs learn a mapping for layout-to-mask interacting with the generator network.
	The main difference between the SPADE module~\cite{park2019semantic} and the proposed ISLA-Norm module lies in how the affine transformation parameters (Eqn.~\ref{eq:isla-gamma} and~\ref{eq:isla-beta}) are computed. The SPADE module is designed for the label-map-to-image synthesis task and thus directly leverages an input ground-truth label map in learning the affine transformation parameters. To compare the designs of SPADE and ISLA-Norm, we conduct experiments in three aspects: 
	\begin{itemize} [leftmargin=*]
		\itemsep0em
		\item \textit{A post-hoc integration} which uses a trained LostGAN-V2 to generate the masks as the input label map to the GauGAN trained with ground-truth label maps in testing. This is to verify that (i) the layout-to-mask generation in LostGANs is meaningful in the sense that the generated masks can be ``dropped in" the GauGANs trained with ground-truth label maps to obtain good image synthesis results, and (ii) the mask-to-image synthesis in LostGANs is sufficiently strong against the counterpart GauGANs when the input masks to the both are the same. Results are reported in Section~\ref{sec:compare_spade}. 
		\item \textit{An end-to-end integration} which uses the mask generation components (i.e., i), ii) and iii) in Section~\ref{sec:isla-norm}), which corresponds to LostGANs-V1, and the SPADE module in GauGANs in training from scratch. This integration is to investigate whether the SPADE module can be used to replace the ISLA-Norm in this straightforward way. However, the training fails in our many tries in the experiments due to the NaN numeric issue. %The generated masks degenerate to and stuck at some trivial solutions, eventually leading to numeric issues.   
		\item \textit{Another end-to-end integration} which uses both the mask generation components and the mask refinement strategy (i.e., iv) in the Section~\ref{sec:isla-norm}), which corresponds to LostGANs-V2, and the SPADE module in GauGANs in training from scratch. This resolves the training issues in the straightforward integration stated above. The resulting model, termed \textit{LostGAN+SPADE}, is slightly worse than the vanilla LostGAN-V2. Results are reported in Section~\ref{sec:compare_spade} with synthesized images shown in Fig.~\ref{fig:model_comparision}. One explanation is that our ISLA-Norm utilizes input bounding boxes to spatially clip the generated and refined masks, while the SPADE module directly uses the generated and refined masks in their entities. 
	\end{itemize}
	
	\subsubsection{The discriminator network}
	As shown in Fig.~\ref{fig:D}, our discriminator consists of three components: a shared ResNet-based feature backbone, an image head classifier and an object head classifier. Detailed network architectures for different image synthesis resolutions are referred to  our Github repository. 
	% \begin{itemize}[leftmargin=*]
	% \itemsep0em
	%     \item \textit{A shared ResNet-based feature backbone} consists of a number of basic ResBlocks~\cite{he2016deep} with BatchNorm modules removed (as commonly adopted in prior arts such as the SNGAN~\cite{miyato2018spectral} and the BigGAN~\cite{brock2018large}). The number of ResBlocks is subject to the target resolution of layout-to-image synthesis. 
	%     \item \textit{The image head classifier} consists of an image-level feature backbone, a global average pooling layer, and an one-output FC layer. The image-level feature backbone includes a number of ResBlocks. The output of the FC layer is a scalar realness score, denoted by $p_{img}$. 
	%     \item \textit{The object head classifier} is a simplified version of R-CNN based object detector~\cite{FasterRCNN,he2017mask} with bounding boxes being given in the input layout. It consists of an object-level feature pyramid, a RoIAlign layer~\cite{he2017mask}, and two FC layers. The one-output FC layer computes a scalar realness score for each object instance, denoted by $p^r_{obj_i}$ ($i\in [1, m]$). To encode object label semantic information, we also compute a label projection score~\cite{miyato2018cgans,brock2018large} for each object instance, denoted by $p^{\ell}_{obj_i}$, which is the inner product between the label embedding (calculated in the same as Eqn.~\ref{eq:embedding}) and the linear projection (using a FC layer) of the RoIAlign feature vector. The overall score of an object instance is the sum: $p_{obj_i}=p^r_{obj_i}+p^{\ell}_{obj_i}$.    
	% \end{itemize} In sum, 
	Denote by $\mathcal{D}(\cdot;\Theta_{\mathcal{D}})$ the discriminator with parameters $\Theta_{\mathcal{D}}$. Given an image $I$ (real or synthesized) and a layout $L$, the discriminator computes a list of scores, 
	\begin{equation}
	(p_{img}, p_{obj_1}, \cdots, p_{obj_m}) = \mathcal{D}(I, L;\Theta_\mathcal{D})
	\end{equation}

	\subsubsection{The loss functions}
	Under the mini-batch based SGD framework, for the generator, the loss function of $\Theta_\mathcal{G}$ is defined by, 
	\begin{align}
		\mathcal{L}(\Theta_\mathcal{G} | \Theta_\mathcal{D}) = - &\underset{(L,I^{syn},I^{gt})\in \mathbb{B}}{\sum}\bigg[P_{\mathcal{D}(I^{syn}, L;\Theta_\mathcal{D})} - \\ 
		\nonumber & ||I^{syn}-I^{gt}||_1 - ||F(I^{syn})-F(I^{gt})||_1\bigg], 
	\end{align}
	where $\mathbb{B}$ represents a mini-batch, $I^{syn}$ and $I^{gt}$ represent a synthesized image (Eqn.~\ref{eq:generator}) and the ground-truth image for the spatial layout $L$,  $P_{\mathcal{D}(I, L;\Theta_\mathcal{D})}=\lambda\cdot p_{img} + \frac{1}{m}\sum_{i=1}^m p_{obj_i}$ with a trade-off parameter $\lambda$ (0.1 used in our experiments), the second term in the right-hand side is the reconstruction loss, and the last term is the perceptual loss~\cite{johnson2016perceptual} which measure L1 difference between features, $F(\cdot)$ of generated image and ground truth images by an ImageNet pretrained network such as the VGG network~\cite{VGG}. Minimizing $\mathcal{L}(\Theta_\mathcal{G} | \Theta_\mathcal{D})$ is trying to fool the discriminator by generating high fidelity images.
	
	For the discriminator, we utilize the hinge version~\cite{HingeLoss1,HingeLoss2} of the standard adversarial loss~\cite{goodfellow2014generative}, 
	\begin{equation}
	l_{t}(I, L) = \begin{cases} \max (0, 1 - p_t);\, & \text{if } I \text{ is a real image} \\
	\max (0, 1 + p_t);\, & \text{if } I \text{ is a fake image} \end{cases}
	\end{equation}
	where $t\in \{img, obj_1, \cdots, obj_m\}$. In the hinge loss, no penalty will occur if the score of a real image (or a real object instance) is greater than or equal to 1, and the score of a fake image (or a fake object instance) is less than or equal to -1. The hinge loss is more aggressive than the real \emph{vs} fake binary classification in the vanilla GAN. The overall loss is,
	\begin{equation}
	l(I, L) = \lambda\cdot l_{img}(I, L) + \frac{1}{m}\sum_{i=1}^m l_{obj_i}(I, L). \label{eq:lossD}
	\end{equation}
	The loss function of $\Theta_{\mathcal{D}}$ is defined by,
	\begin{equation}
	\mathcal{L}(\Theta_\mathcal{D} | \Theta_\mathcal{G}) = \underset{(L,I^{syn},I^{gt})\in \mathbb{B}}{\sum}\Big[l(I^{gt}, L)+l(I^{syn}, L)\Big],  
	\end{equation}
	where $p(I, L)$ represents both the real and fake (synthesized by the generator) data. Minimizing $\mathcal{L}(\Theta_\mathcal{D} | \Theta_\mathcal{G})$ is trying to tell apart the real and fake images. 
	
	\subsubsection{Layout semi-supervised training of LostGANs}
	Conditional image synthesis is a data and annotation hungry task. For layout-to-image synthesis, it is interesting to investigate the paradigm of learning with less labels (i.e., layouts). We present a straightforward two-stage training procedure. 
	
	Denote by $D_s$ and $D_u$ the image dataset with and without object bounding boxes annotated respectively. We first train a Faster-RCNN~\cite{FasterRCNN} object detector using $D_s$. Then, we apply the trained Faster-RCNN detector in $D_u$. In terms of how to leverage the detection results in $D_u$, the most obvious way is to use the bounding boxes of detected objects whose probabilities are greater than a predefined threshold, e.g., $0.5$. This gives us reasonably results. 
	
	To account for the uncertainty of Faster-RCNN detection results, we present a detection score re-weighing method which uses the detection probability of a declared object bounding box by the Faster-RCNN detector in the loss of the discriminator network. Let $L=(\ell_i, \{bbox_i, p_i\}_{i=1}^m)$ the layout for an image $I\in D_u$ based on the detection results, where $p_i\geq \tau$ (e.g., $\tau=0.5$). Eqn.~\ref{eq:lossD} is rewritten as, 
	\begin{equation}
	l(I, L) = \lambda\cdot l_{img}(I, L) + \frac{1}{m}\sum_{i=1}^m p_i\cdot l_{obj_i}(I, L), \label{eq:lossD1}
	\end{equation}
	which leads to comparable performance to the fully-supervised LostGAN when only half of the images in the COCO-Stuff dataset use annotated bounding boxes (see results analyses in Section~\ref{sec:semi}).

	\begin{table*}[t]
		\centering
		\caption{Quantitative comparisons using the Inception Score (IS, higher is better, illustrated by $\uparrow$), FID (lower is better, illustrated by $\downarrow$) and Diversity Score (DS, higher is better, illustrated by $\uparrow$) evaluation metrics in the COCO-Stuff~\cite{caesar2018coco} and VG~\cite{krishna2017visual} datasets. See text for details.}
		\label{tab:eval_results}
		% \resizebox{0.9\textwidth}{!}{
		\begin{tabular}{r|rr|rr|rr}
			\hline
			\multirow{2}{*}{\textbf{Methods}} & \multicolumn{2}{c|}{\textbf{IS}$\uparrow$} & \multicolumn{2}{c|}{\textbf{FID}$\downarrow$} & \multicolumn{2}{c}{\textbf{DS}$\uparrow$} \\
			& \textbf{COCO} & \textbf{VG} & \textbf{COCO} & \textbf{VG} & \textbf{COCO} & \textbf{VG} \\ \hline
			Real Images (64$\times$64) & 16.30 $\pm$ 0.40 & 13.90 $\pm$ 0.50 & - & - & - & - \\
			Real Images (128$\times$128) & 22.30 $\pm$ 0.50 & 20.50 $\pm$ 1.50 & - & - & - & - \\
			Real Images (256$\times$256) & 28.10 $\pm$ 1.60 & 28.60 $\pm$ 1.20 & - & - & - & - \\
			Real Images (512$\times$512) & 34.50 $\pm$ 1.70 & 34.20 $\pm$ 1.10 & - & - & - & - \\
			\hline
			pix2pix~\cite{isola2017image} 64$\times$64 & 3.50 $\pm$ 0.10 & 2.70 $\pm$ 0.02 &121.97 &142.86 & 0 & 0 \\
			% BicycleGAN~\cite{bicyclegan} 64$\times$64 & 6.40 $\pm$ 0.10 & 5.60 $\pm$ 0.10 &82.48 &80.35 & 0.33 $\pm$ 0.14 & 0.35 $\pm$ 0.12 \\
			% pix2pixHD~\cite{wang2018high} 64$\times$64 & 4.70 $\pm$ 0.10 & 4.50 $\pm$ 0.10 &121.21 &83.12 & 0.24 $\pm$ 0.10 & 0.26 $\pm$ 0.12 \\
			sg2im (GT Layout)~\cite{johnson2018image} 64$\times$64 & 7.30 $\pm$ 0.10 & 6.30 $\pm$ 0.20 &67.96 &74.61 & 0.02 $\pm$ 0.01 & 0.15 $\pm$ 0.12 \\
			% GauGAN~\cite{park2019semantic} 64$\times$64 & 8.30 $\pm$ 0.20 & 6.40 $\pm$ 0.20 &49.27 &47.74 & 0.18 $\pm$ 0.06 & 0.18 $\pm$ 0.11 \\
			Layout2Im~\cite{zhao2020layout2image} 64$\times$64& 9.10 $\pm$ 0.10 & 8.10 $\pm$ 0.10 & 44.19 & 39.68 & 0.15 $\pm$ 0.06 & 0.17 $\pm$ 0.09 \\ 
			Layout2Im + OWA~\cite{zhao2020layout2image} 64$\times$64& 9.70 $\pm$ 0.10 & 8.00 $\pm$ 0.20 & 40.19 & 33.54 & 0.09 $\pm$ 0.05 & 0.09 $\pm$ 0.11 \\ 
			\textbf{Our LostGAN-V1}~\cite{LostGAN} 64$\times$64 & \textbf{9.80 $\pm$ 0.20 }& \textbf{8.70 $\pm$ 0.40} & \textbf{34.31} & \textbf{34.75} & \textbf{0.35 $\pm$ 0.09} & \textbf{0.34 $\pm$ 0.10} \\ \hline
			Grid2Im~\cite{ashual2019specifying} (GT Layout) 128$\times$128 & 11.22 $\pm$ 0.15  & - & 63.44 &- & 0.28 $\pm$ 0.11 & - \\ 
			Our LostGAN-V1~\cite{LostGAN} 128$\times$128 & 13.80 $\pm$ 0.40 & \textbf{11.10 $\pm$ 0.60} & 29.65 & 29.36 & 0.40 $\pm$ 0.09 & \textbf{0.43 $\pm$ 0.09 }\\ 
			\textbf{Our LostGAN-V2 }128$\times$128 & \textbf{14.21 $\pm$ 0.40} & {10.71 $\pm$ 0.26} & \textbf{24.76} & \textbf{29.00} & \textbf{0.45 $\pm$ 0.09} & 0.42 $\pm$ 0.09\\\hline
			Grid2Im~\cite{ashual2019specifying} (GT Layout) 256$\times$256 &15.23 $\pm$ 0.11 & - & 65.95 & -& 0.34 $\pm$ 0.13 & - \\ 
			\textbf{Our LostGAN-V2} 256$\times$256 & \textbf{18.01 $\pm$ 0.50} & 14.10 $\pm$ 0.38 & \textbf{42.55} & 47.62 & \textbf{0.55 $\pm$ 0.09} & 0.53 $\pm$ 0.09 \\ \hline
			%Our Mask + GauGAN~\cite{park2019semantic} 256$\times$256 & 19.35 $\pm$ 0.73 & - & 41.11 & - & 0.38 $\pm$ 0.12 & - \\
			% \hline %move to ablation study 
			{Our LostGAN-V2} 512$\times$512 & 17.55 $\pm$ 0.23 & {14.42 $\pm$ 0.46} & 51.99 & {52.73} & {0.65 $\pm$ 0.11} & {0.61 $\pm$ 0.09} \\
			% \textbf{LostGAN-V2} 256to512 & \textbf{18.10 $\pm$ 0.45} & 14.30 $\pm$ 0.38 & \textbf{45.37} & 54.55 & 0.61 $\pm$ 0.10 & 0.59 $\pm$ 0.10 \\
			\hline
		\end{tabular}
		% }
	\end{table*}
	
	\subsubsection{Implementation details}
	% The proposed model follows \cite{miyato2018cgans,zhang2018self,brock2018large}, using the ResNet\cite{he2016deep} GAN structure.
	In training, we follow the practice used in~\cite{miyato2018cgans,zhang2018self,brock2018large}. Synchronized BatchNorm~\cite{ioffe2015batch}, where batch statistics for feature standardization are computed over all devices, is adopted in our ISLA-Norm. The Spectral Normalization \cite{miyato2018spectral} of model parameters is also applied in both the Generator and the Discriminator to stabilize training. Parameters of the Generator and the Discriminator are initialized using the Orthogonal Initialization method~\cite{saxe2013exact}. The Adam optimizer \cite{kingma2014adam} is used with $\beta_1 = 0$ and $\beta_2 = 0.999$. The learning rate is set constant $10^{-4}$ for both the Generator and the Discriminator. We use a batch size of 128 based on our computing resource. %Training our LostGAN takes about 2-3 days, \eg, either for the models generating $128\times128$ images on 2 NVIDIA V100 GPUs, or for the models synthesizing $256\times256$ images on 4 NVIDIA V100 GPUs.

	\section{Experiments} \label{sec:experiments}
	We test our LostGANs in the COCO-Stuff dataset \cite{caesar2018coco} and the Visual Genome (VG) dataset \cite{krishna2017visual}. We evaluate LostGAN-V1 at two resolutions (64$\times$64 and 128$\times$128) and LostGAN-V2 at three resolutions (128$\times$128, 256$\times$256 and $512\times512$). We evaluate the Semi-LostGAN-V2 at the resolution of 128$\times$128. Our LostGAN-V2 obtains state-of-the-art performance. 
	% \textbf{Our source code and pretrained models have been made publicly available at} \url{https://github.com/iVMCL/LostGANs}. 

	\textbf{Datasets.}
	The \textbf{COCO-Stuff} 2017 \cite{caesar2018coco} augments the COCO dataset with pixel-level stuff annotations. The annotation contains 80 \textit{thing} classes (person, car, \textit{etc.}) and 91 \textit{stuff} classes (sky, road, \textit{etc.})
	Following settings of \cite{johnson2018image}, objects covering less than 2\% of the image area are ignored, and we use images with 3 to 8 objects. 
	For the \textbf{Visual Genome} (VG) dataset \cite{krishna2017visual}, we follow the settings of \cite{johnson2018image} to  remove small and infrequent objects, which results in 62,565 images for training, 5,506 images for validation and 5,088 images for testing, with 3 to 30 objects from 178 categories in each image.

	\textbf{Methods in comparison.}
	We compare with four prior arts:
	\textit{i) The pix2pix} method~\cite{isola2017image} learns to map images between two domains. We reuse the pix2pix results reported in the Layout2Im~\cite{zhao2020layout2image} in our comparisons, where a pix2pix model is trained to synthesize images from a feature map learned to encode the layout. The number of channels of the feature map is the number of categories (\eg, 171 in COCO-Stuff). 
	\textit{ii) The scene graph to image (sg2im)} method~\cite{johnson2018image} synthesizes images from input scene graphs with an intermediate scene-graph-to-layout module. We compare with sg2im using the ground-truth (GT) layouts. 
	\textit{iii) The Layout2Im} method~\cite{zhao2020layout2image} is the first to synthesize images directly from input layouts. These three methods have only been evaluated at the resolution of 64$\times$ 64. \textit{iv) The Grid2Im method}~\cite{ashual2019specifying} extends the sg2im method, which has been tested at two resolutions, 128$\times$128 and 256$\times$256, in the COCO-Stuff dataset only since ground-truth masks are needed in training. We also compare with Grid2Im using the GT layouts. %Fig.~\ref{fig:model_comparision} and~\ref{fig:model_comparision_vg} show examples of synthesized images by different methods. 

	\textbf{Evaluation metrics.}
	It remains a challenging problem to automatically evaluated image synthesis in general. For the layout-to-image synthesis, we adopt four state-of-the-art metrics and test a new one specifically reflecting the layout quality as follows. 
	
	The \textit{Inception Score} (IS)~\cite{salimans2016improved} uses an Inception V3 network pretrained on the ImageNet-1000 classification benchmark and computes a score (statistics) of the network's outputs with $N$ synthesized images $I_i$'s of a generator model $\mathcal{G}$. The IS aims to capture two desirable qualities of image synthesis: Synthesized image should contain clear and meaningful objects (subject to the ImageNet-1000 training datasets), and diverse images from all the different categories in ImageNet should be observed in synthesized images. So, \textit{the larger the IS is, the better a generator model is}. Multiple runs are usually used to calculate the mean$\pm$std evaluation (\eg, 5 runs are typically used). The IS does not leverage the statistics of real images. 
	
	The \textit{Fr\`echet Inception Distance} (FID)~\cite{FID} has been proposed to improve IS by incorporating statistics from real images. It also uses an ImageNet-pretrained Inception V3 network and computes the  Fr\`echet distance~\cite{FrechetDistance} between two Gaussian distributions fitted to synthesizes images and real images respectively. \textit{The lower the FID is, the better a generator model is}. Both the IS and FID do not explicitly measure the quality of one-to-many mapping in layout-to-image synthesis. 
	
	The \textit{Diversity Score} (DS) aims to compare the perceptual similarity in a DNN feature space between two images, $I_1$ and $I_2$, randomly generated from the same layout. We adopt the LPIPS metric~\cite{zhang2018unreasonable} in computing the DS. \textit{The higher the DS is, the better a generator model is}. 
	
	The \textit{Classification Accuracy Score} (CAS)~\cite{ravuri2019classification}. One long-term goal of generative learning in practice is to leverage synthesized images in training discriminative models. The CAS aims to verify how well a classification model trained only on synthesized images can perform on real testing images. So, \textit{the higher the CAS is, the better a generator model is.} In contrast to the CAS, the classification accuracy metric used in the Layout2Im~\cite{zhao2020layout2image} is based on models trained with real image and tested on synthesized images, which may overlook the diversity of synthesized images. 
	
	The \textit{Object Detection Average Precision (DAP)}. To evaluate the quality of a synthesized image in its entirety and to reflect the quality of the layout of a synthesized image, we first train a Faster-RCNN~\cite{FasterRCNN} using the training data, and then evaluate the detection performance of the Faster-RCNN detector in the synthesized images generated using the layouts in the validation dataset.   
	
	\subsection{Results}
	% We first present the quantitative results and analyses. Table~\ref{tab:eval_results} and Table~\ref{tab:classification_score} summarize comparisons in terms of the our metrics in the two datasets. 
	
	\subsubsection{Overall synthesis quality based on IS, FID and DS}
\begin{figure*}[t] 
		\centering
		\includegraphics[width=0.99\linewidth]{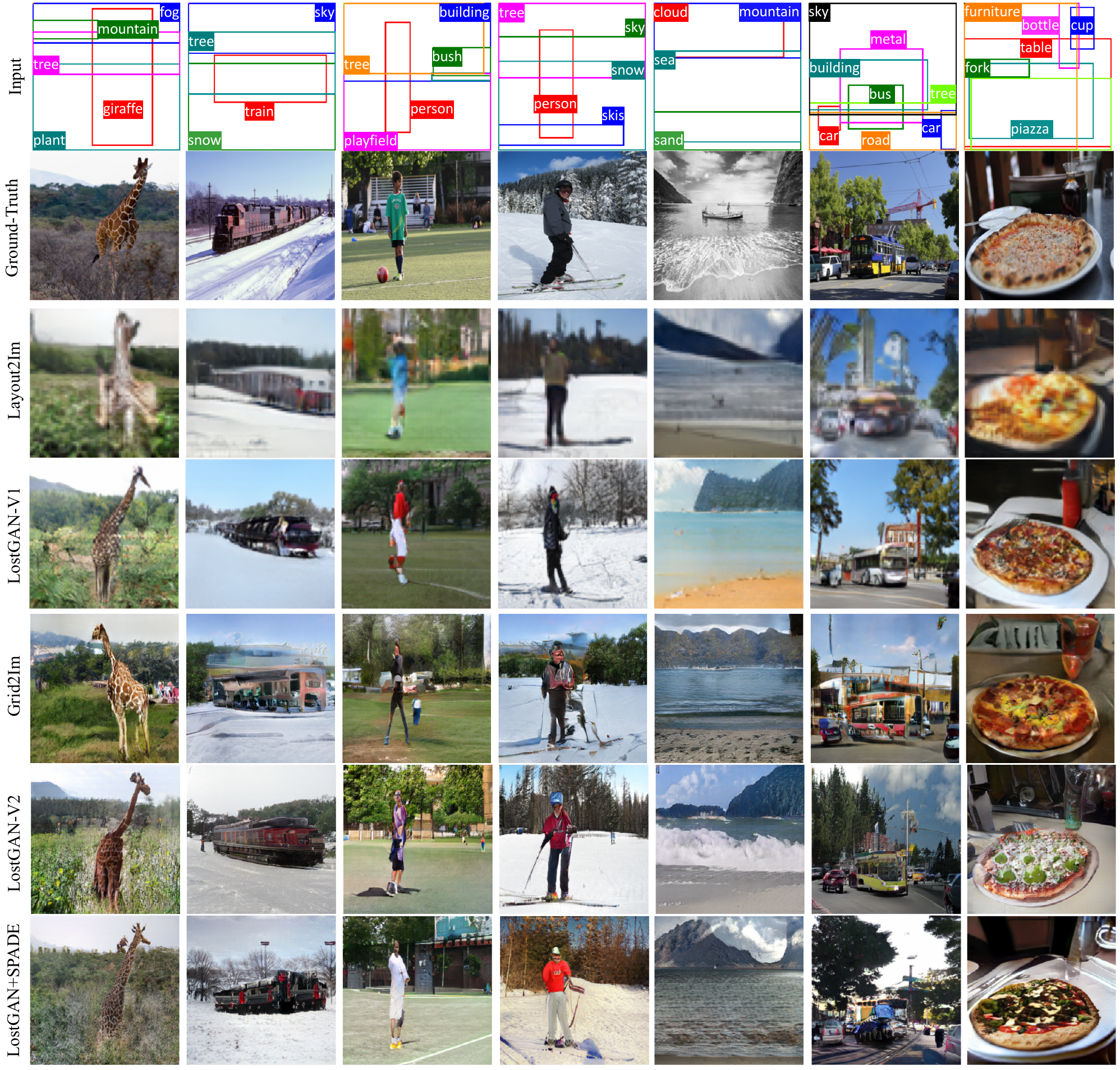}
		\caption{Synthesized images from given layouts in COCO-Stuff by different models. From the top to the bottom: Input layout, Ground-truth image, Layout2Im~\cite{zhao2020layout2image} $64\times 64$, our LostGAN-V1~\cite{LostGAN} $128\times 128$,  Grid2Im~\cite{ashual2019specifying}  $256\times 256$, our LostGAN-V2 $256\times 256$, and LostGAN+SPADE~\cite{park2019semantic} (end-to-end integration, the third method in Section~\ref{sec:compare_spade}) $256\times 256$.}
		\label{fig:model_comparision}
	\end{figure*}

\begin{figure}[!ht] 
		\centering
		\includegraphics[width=0.98\linewidth]{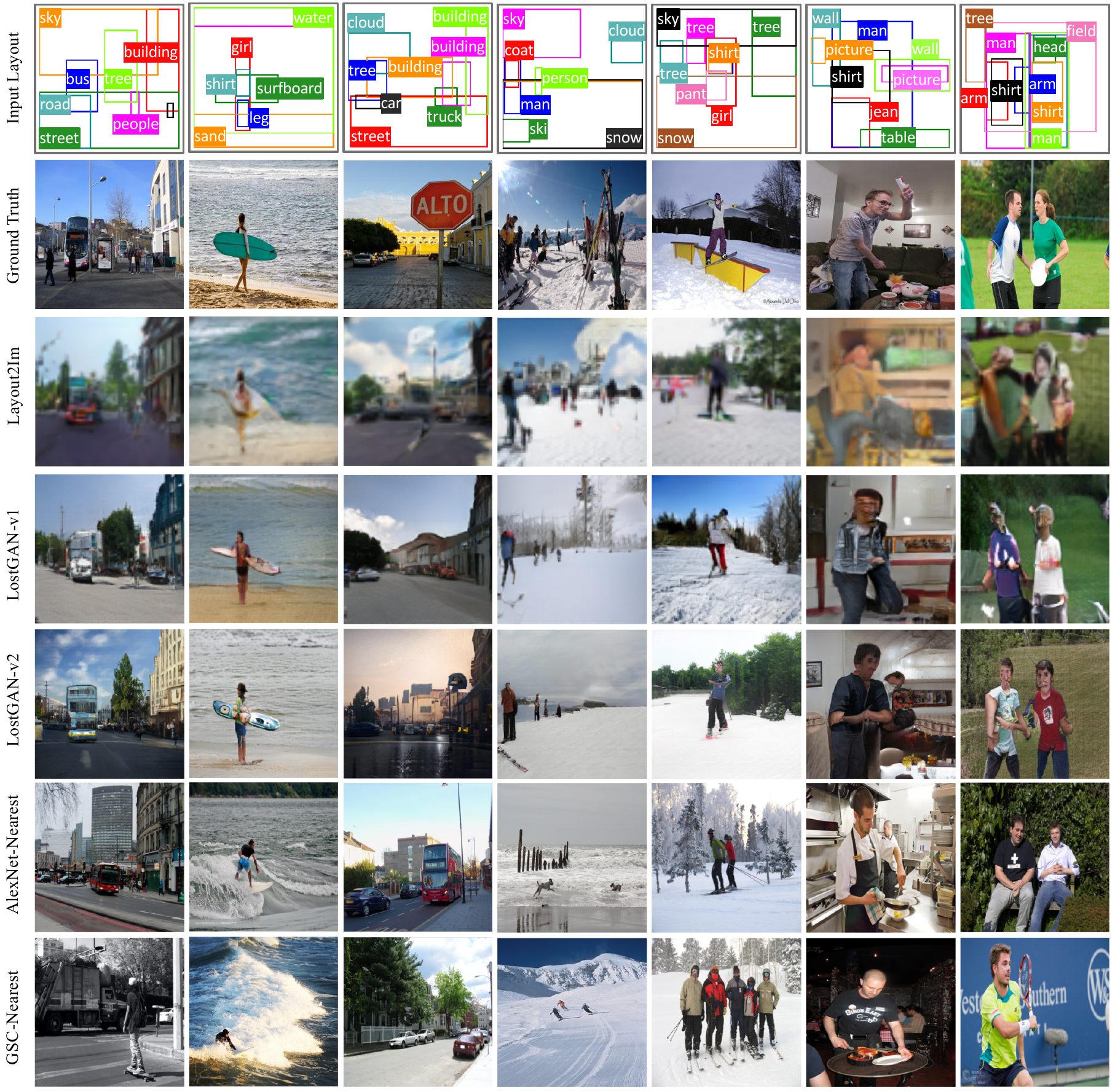}
		\caption{Synthesized images in VG by different models: Layout2Im~\cite{zhao2020layout2image} $64\times 64$, our LostGAN-V1 $128\times 128$, our LostGAN-V2 $256\times 256$. The last two rows show the nearest neighbors of the synthesized images by our LostGAN-V2 in the VG training dataset using the AlexNet-pool5 feature ~\cite{krizhevsky2012imagenet} and the GSC metric~\cite{bansal2019shapes} (Sec.~\ref{sec:nn}).}
		\label{fig:model_comparision_vg}
	\end{figure}

	Table~\ref{tab:eval_results} summarizes the comparisons. 
	Fig.~\ref{fig:model_comparision} and Fig~\ref{fig:model_comparision_vg} show images synthesized by different models from the same layout in  COCO-Stuff and VG respectively. The input layouts are quite complex. Our LostGAN-V2 can generate visually more appealing images with more recognizable objects that are consistent with input layouts at resolution 256$\times$256. 
	We analyze the quantitative results as follows. 
	
	\textbf{At the resolution of $64\times 64$}. Our LostGAN-V1 obtains the best performance in comparison. It obtains slightly better Inception Score in both datasets and FID in the VG dataset than the Layout2Im. It obtains significantly better FID in the COCO-Stuff dataset (by more than 5 points reduction) and DS in both datasets. The diversity score of our LostGAN-V1 outperforms the Layout2Im by relative 288.9\% and 277.8\% in the two datasets respectively. There are a few other methods tested at the resolution of $64\times 64$ in the Layout2Im~\cite{zhao2020layout2image}, including the pix2pixHD~\cite{wang2018high}, BicycleGAN~\cite{bicyclegan} and GauGAN~\cite{park2019semantic}, which are outperformed by the Layout2Im method and thus not included here for the clarity of the table.  
	
	\textbf{At the resolution of $128\times 128$}. Our LostGAN-V1 obtains better results than the Grid2Im method in the COCO-Stuff dataset, especially by more than $33\%$ reduction in FID and by relative 42.9\% increase in DS. Our LostGAN-V2 further improves the results, except for the IS and DS in the VG dataset. The decrease of IS and DS may be caused by the factors as follows.  
	
	\textit{Remarks.} We observed that the VG dataset includes more diverse object configurations (\eg, bounding boxes may severely overlap in an image such as those for people, cloth and pants). In general, the bounding box annotations in the VG dataset are of lower quality than those in the COCO-Stuff dataset (\eg, they may have significant offsets for certain object instances). Those factors may affect the layout-to-mask component, especially the module of predicting masks from feature maps in the generator, which we think is the reason of LostGAN-V1 slightly outperforming LostGAN-V2 in the VG dataset. Similarly, Layout2Im+OWA~\cite{zhao2020layout2image} suffers a slight drop of performance in the VG dataset after introducing an object-wise attention mechanism to model shape of different objects. Considering those, we only test our LostGAN-V2 at higher resolutions than $128\times 128$. 
	
	\textbf{At the resolution of $256\times 256$}. Our LostGAN-V2 also obtains better results than the Grid2Im method by more than 2\% increase in IS, 23\% reduction in FID and relative 61.8\% increase in DS in the COCO-Stuff dataset. 
	
\begin{figure}[t]
		\centering
		\includegraphics[width=0.98\linewidth]{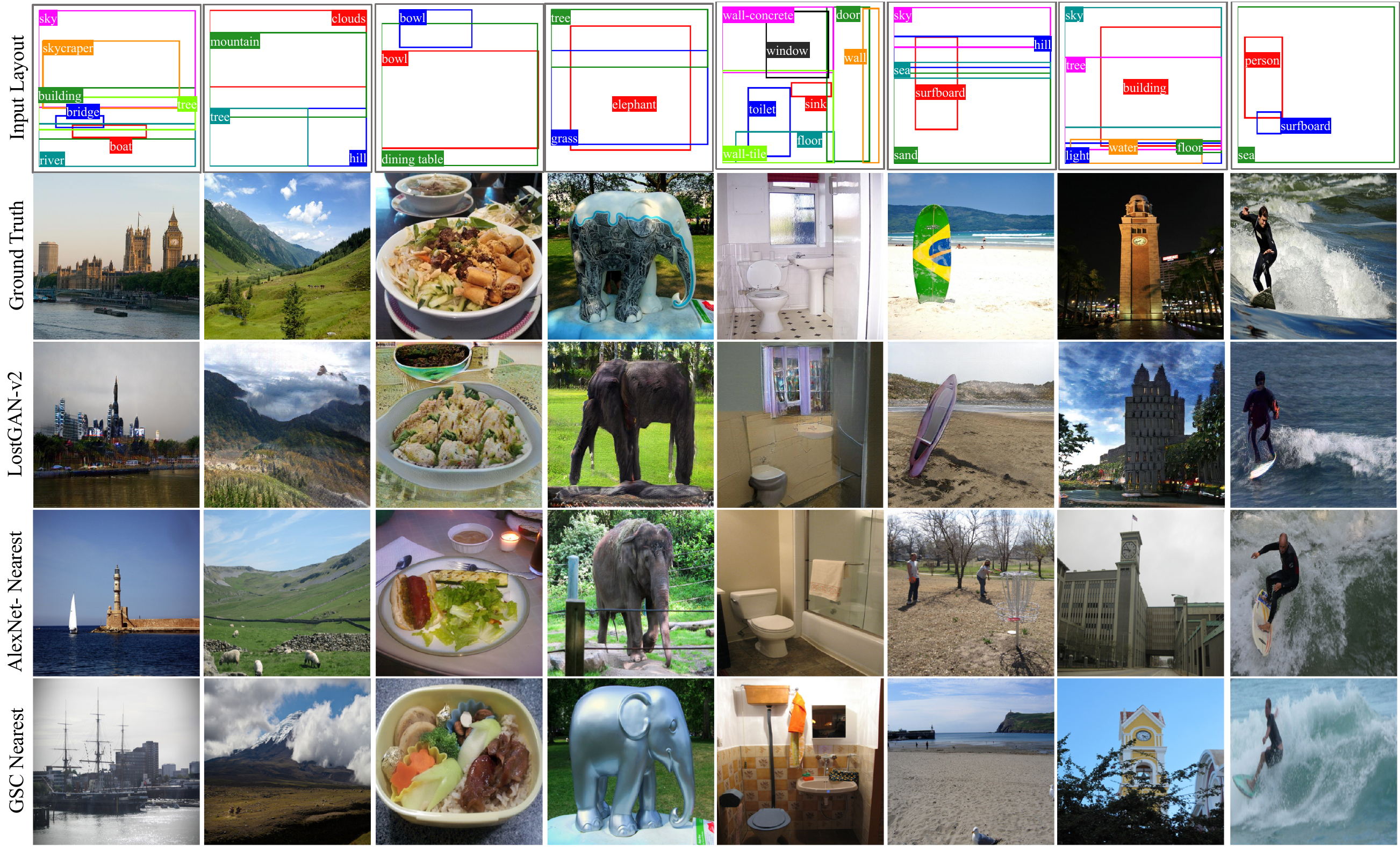}
		\caption{Some selected examples of synthesized images at the resolution of $512\times512$ in COCO-Stuff by our LostGAN-V2. The last two rows show the nearest neighbors of the synthesized images by our LostGAN-V2 in the CoCo-Stuff training dataset using the AlexNet-pool5 feature ~\cite{krizhevsky2012imagenet} and the GSC metric~\cite{bansal2019shapes} (Sec.~\ref{sec:nn}).}
		\label{fig:coco512}
	\end{figure}

	\textbf{At the resolution of $512\times 512$}. There is no results from other baselines. Our LostGAN-V2 obtains better DS than that at the resolution of $256\times 256$. However, our LostGAN-V2 obtains slightly worse results than those obtained at the resolution of $256\times 256$ in terms of IS and FID. This phenomenon has been also observed in the BigGAN~\cite{brock2018large}, which indicates, on the one hand, that more research are entailed to improve the quality of high resolution image synthesis, and on the other hand, that the models (Inception V3 pretrained in ImageNet at the resolution of $300\times 300$) used in computing IS and FID may need to change. 
	Fig.~\ref{fig:coco512} shows some selected examples of synthesized images. We observed that it is more difficult to generate realistic looking images at the resolution of $512\times 512$.
	
	%We perform a across-resolution comparison: we upsample the synthesized images at the resolution of $256\times 256$ to the resolution of $512\times 512$ (\ie, ``256to512" in Table~\ref{tab:eval_results}) and similar observations retain.  

	\subsubsection{Object synthesis quality based on CAS and DAP}
	Table~\ref{tab:classification_score} summarizes comparisons of CAS. To compare the CAS, we train the ResNet-101~\cite{he2016deep} on cropped and resized objects at a resolution of $32\times 32$ from generated images (five samples generated for each layout in the testing set) and evaluate the trained model on objects cropped and resized from real testing images. We follow the widely used settings of ResNet-101 on the CIFAR-10/100 (with images at the resolution $32\times 32$). We train a 171-category classification ResNet-101 in the COCO-Stuff dataset and a 178-category ResNet-101 in the VG dataset. For synthesized images at the three resolutions, our LostGANs obtain the best accuracy, often by large margin. These results are aligned with the higher DS results consistently obtained by our methods. Hopefully, with more research in the future work, we will be able to generate high-fidelity and high-resolution images from reconfigurable layouts and styles to faciliate more powerful discriminative learning, especially for handling some long-tail or corner situations.    
	
	\begin{table}[!ht]
		\centering
		\caption{Comparisons of the CAS. See text for details.}
		\label{tab:classification_score}
		\begin{tabular}{r|cc}
			\hline
			\multirow{2}{*}{\textbf{Methods}} & \multicolumn{2}{c}{\textbf{CAS}$\uparrow$} \\
			& COCO & VG \\
			\hline
			Layout2Im~\cite{zhao2020layout2image} 64x64  & 27.32 & 23.25\\
			\textbf{Our LostGAN-V1}~\cite{LostGAN} 64x64 & \textbf{28.81} & \textbf{27.50} \\ \hline
			Grid2Im~\cite{ashual2019specifying} 128x128 & 25.89 & - \\
			Our LostGAN-V1~\cite{LostGAN} 128x128 & 30.68 & 28.85 \\
			\textbf{Our LostGAN-V2} 128x128 & \textbf{31.98} & \textbf{29.35} \\ \hline
			Grid2Im~\cite{ashual2019specifying}  256x256 & 20.54 & - \\
			\textbf{Our LostGAN-V2} 256x256 & \textbf{30.33} & \textbf{28.81} \\ \hline
			% 30.49
			Real Images & 51.04 & 48.07 \\
			\hline
		\end{tabular}
	\end{table} 
	\begin{table}[!h]
		\centering
		\caption{Comparisons of the detection AP (DAP$\uparrow$). See text for details.}
		\label{tab:DAP}
		\begin{tabular}{ c|c|c|c } 
			\hline 
			\textbf{Testing} & AP$^{bb}$ & AP$_{50}^{bb}$ & AP$_{75}^{bb}$ \\ \hline 
			Gird2Im~\cite{ashual2019specifying} 256x256 & 32.3 & 58.5 & 30.9 \\ \hline
			\textbf{Our LostGAN-V2} 256x256 & \textbf{34.9} & \textbf{60.8} & \textbf{34.4} \\\hline 
			Real Images & 44.7 & 69.5 & 47.5 \\
			\hline 
		\end{tabular}
	\end{table}
	
	Table~\ref{tab:DAP} summarizes the detection Average Precision (AP) comparisons between our LostGAN-V2 and the prior art, the Grid2Im method. We report the standard COCO metrics of average precision, AP$^{bb}_{50}$ using the intersection-over-union (IOU) threshold $0.5$, and AP$^{bb}_{75}$ using the IOU threshold $0.75$, and AP$^{bb}$ with IOU thresholds accumulated from $0.5$ to $0.9$. Our LostGAN-V2 achieves 2.6\% absolute increase of the AP.

	\subsubsection{Real image nearest neighbors of generated images}\label{sec:nn}
	To further show the fidelity of generated images compared with real images at an exemplar level in addition to the FID at a distribution level, and to check if the model overfits the training data, we compute two types of nearest neighbors in the training dataset for an image synthesized using an input layout in the validation dataset: (i) \textit{The AlexNet-Nearest} for an synthesized image: It is computed based on the cosine similarities between the synthesized image and images in the training dataset. Images are represented by the {\tt pool5} layer features in the ImageNet pretrained AlexNet~\cite{krizhevsky2012imagenet}, which are $256\times 7\times 7$ feature maps for $256\times 256$ input images. This will capture both the appearance similarity and the coarse structural similarity between synthesized images and training images. (ii) \textit{The GSC-Nearest} for the input layout: The Global Scene Consistency (GSC)~\cite{bansal2019shapes} metric is used, which measure both the distance of normalized histogram of labels and pixel-to-pixel overlap between query and target layouts, to find the nearest neighbors of the input layout map. Results are shown in Fig.~\ref{fig:model_comparision_vg}, Fig.~\ref{fig:coco512} and Fig.~\ref{fig:various_outs} (together with other aspects of the proposed method), from which we can see the nearest neighbors are semantically meaningfully aligned with the generated images and the synthesized images are visually different from the nearest neighbor in terms of appearance and structure. This also supports the results that our LostGANs outperform other methods in terms of FID (Table~\ref{tab:eval_results}). 
	
	% utilize an ImageNet-pretrained VGG-16 network~\cite{VGG} and use the ``fc7" layer output as the feature representation. We find the top-5 real image nearest neighbors for a synthesized image. We also use the Global Scene Consistency \cite{bansal2019shapes}, which measure both the distance of normalized histogram of labels and pixel-to-pixel overlap between query and target layouts, to find the nearest neighbors of the input layout map. Fig.~\ref{fig:vgg_nearest_neighbor} shows three examples, from which we can see the nearest neighbors are semantically meaningfully aligned with the generated images. This also supports the results that our LostGANs outperform other methods in terms of FID (Table~\ref{tab:eval_results}). 

	\begin{figure}[ht]
		\centering
		\includegraphics[width=0.98\linewidth]{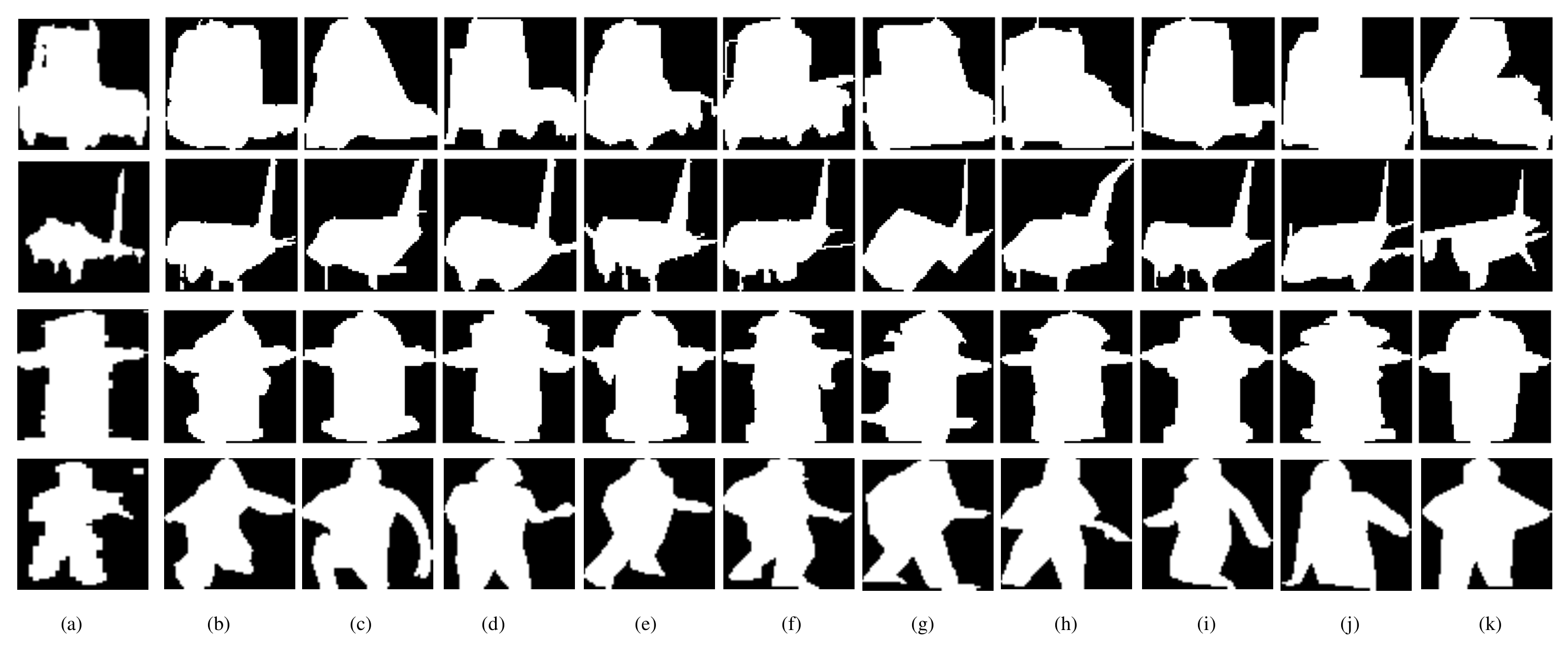}
		\caption{Examples of learned masks and their nearest neighbors in the ground-truth masks in the COCO-Stuff training dataset: truck, airplane, hydrant and person (from top to bottom). (a) Masks learned by our LostGAN-V2 $256\times 256$ and (b-k) top-$10$ nearest neighbors.  All masks are cropped and resized to the resolution of $32\times 32$. See text for details.}
		\label{fig:nearest_mask}
	\end{figure}
	\begin{table}[ht]
		\centering
		\caption{mIoU between masks and their nearest neighbor in ground truth masks.}
		\label{tab:mask_miou}
		\resizebox{0.48\textwidth}{!}{
			\begin{tabular}{|c|c|c|c|c|c|c|}
				\hline
				person & car & plane & bus & train & truck & boat \\ \hline
				53.8 & 66.5 & 58.0 & 75.0 & 70.8 & 66.1 & 63.1 \\ \hline
				zebra & hydrant & pizza & elephant & laptop & bench & \textbf{mean} \\ \hline
				66.9 & 59.2 & 77.7 & 62.3 & 57.0 & 62.8 & 56.5 \\ \hline
			\end{tabular}
		}
	\end{table}
	\subsubsection{Evaluation of the weakly-supervised learning of layout-to-mask generation in LostGANs}
	To investigate the quality of learned masks, we resort to the intersection-over-union (IoU) metric used in object semantic segmentation. We measure the IoU performance in the COCO-Stuff training dataset. We first crop masks for each category and then resize all the masks to the same resolution of $32\times 32$. After training the LostGAN-V2 $256\times 256$, we run inference on each layout in the training dataset (one run is used for simplicity) and obtain the learned masks. We then crop and resize object masks in the same way as done for the ground-truth object masks. For each learned object mask, we retrieve the top-$k$ nearest neighbors in terms of mask IoU in the set of ground-truth object masks.  Fig.~\ref{fig:nearest_mask} shows four examples with the top-$10$ nearest neighbors. Table~\ref{tab:mask_miou} shows the mean IoUs for 13 selected object categories which have reasonably high IoUs. Section~\ref{sec:layout2mask} shows qualitative analyses of the layout-to-mask module.

	\subsection{Qualitative analyses}
	\subsubsection{Controllability and reconfigurability of style and layout}
	We show more examples of layout and style control in our LostGAN-V2 as follows, in addition to Fig.~\ref{fig:teaser}. 
	
\begin{figure}[!ht]
		\centering
		\includegraphics[width=0.95\linewidth]{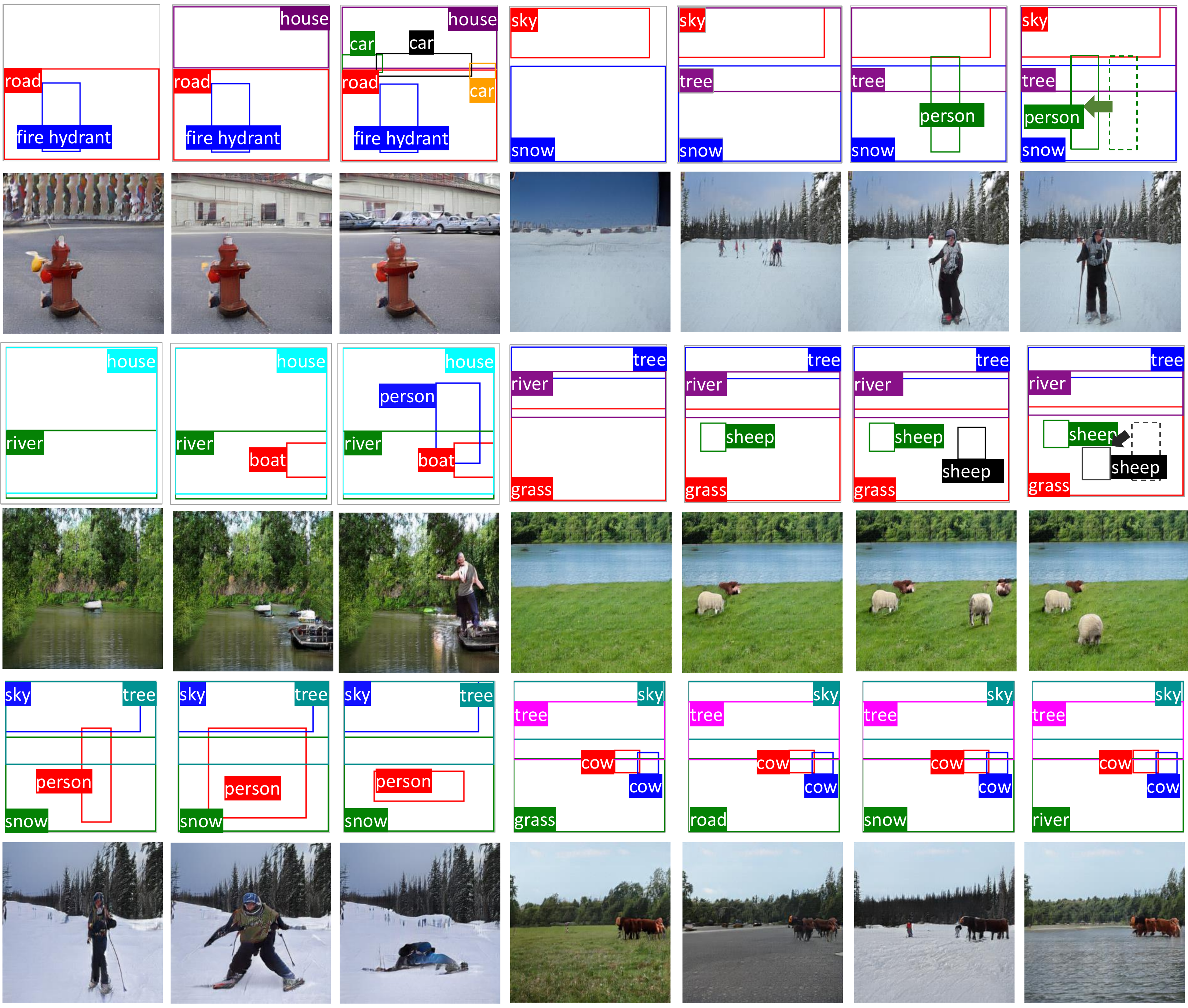}
		\caption{\textbf{Layout Control} in our LostGAN-V2: image synthesis results by adding new objects, changing the spatial position, the size, the aspect ratio or the category label of a bounding box in a layout. Best viewed in magnification and color.}
		\label{fig:reconfigrable_layout}
	\end{figure}

	\textbf{Layout controllability} is demonstrated by adding object to, or moving a bounding box in a layout. As shown in Fig.~\ref{fig:reconfigrable_layout}, when adding extra objects or moving the bounding box of one instance, our model can generate reasonable objects at the desired position while keeping existing objects unchanged as we keep the input style vectors of existing objects fixed. When moving the bounding box of an existing object, the style of generated object at the new position also is kept consistent, \eg, in the top-right of Fig.~\ref{fig:reconfigrable_layout}, the person bounding box is moved, while the style of the synthesized person is retained such as the pose and the color of clothes. 
	
	\begin{figure*}
		\centering
		\includegraphics[width=0.99\linewidth]{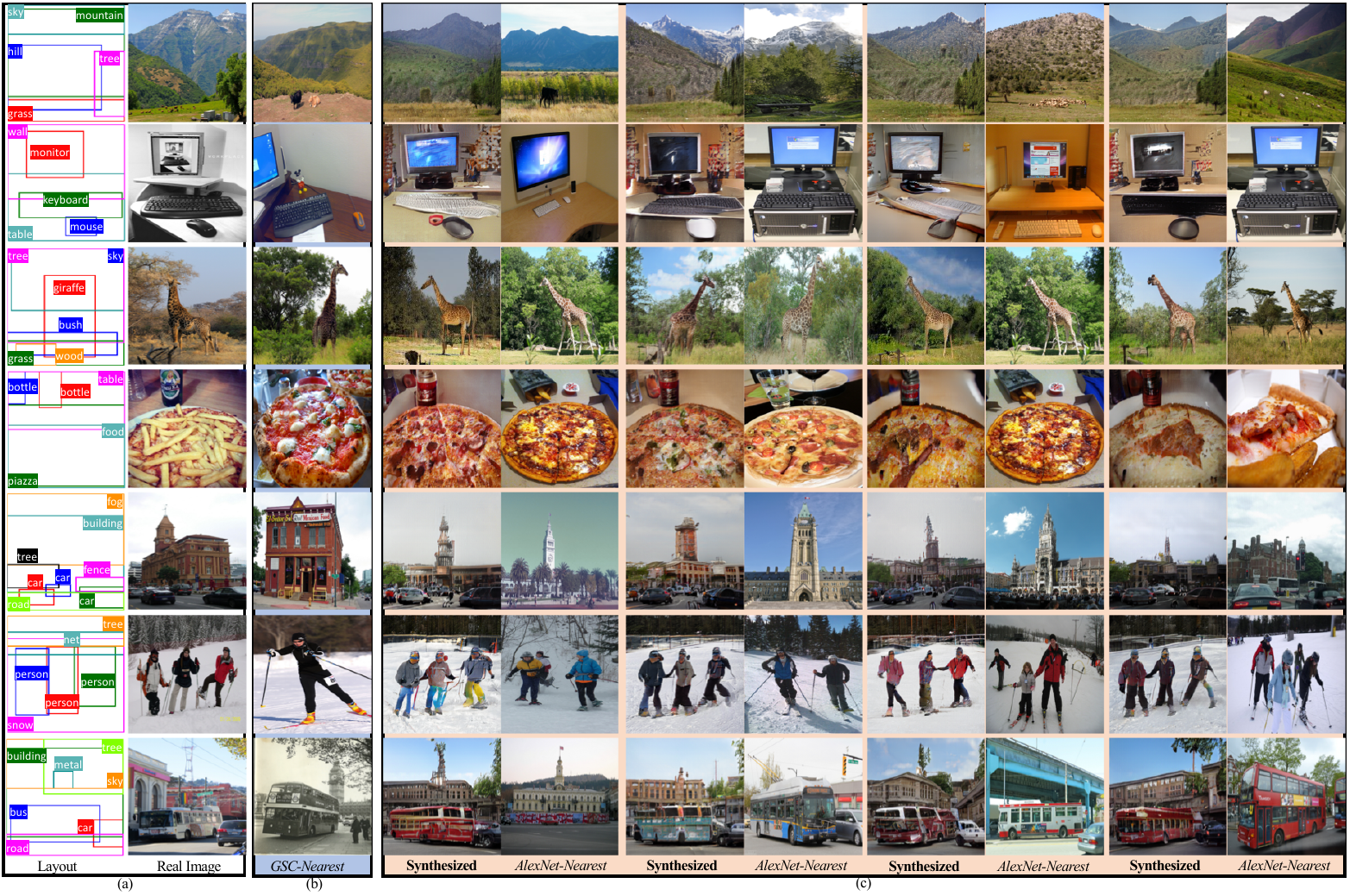}
		\caption{\textbf{Style Control} w.r.t. $(z_{img}, Z_{obj})$ in our LostGAN-V2 $256\times 256$: multiple images synthesized using the same layout with different styles, $(z_{img}, Z_{obj})$'s. We also show the nearest neighbors of the synthesized images by our LostGAN-V2 in the CoCo-Stuff training dataset using the AlexNet-pool5 feature ~\cite{krizhevsky2012imagenet} and the GSC metric~\cite{bansal2019shapes} (Sec.~\ref{sec:nn}). (a) Layout and GT real image, (b) the GSC-Nearest based on the input layout, and (c) Synthesized images by our LostGAN-V2 $256\times 256$ and their AlexNet-Nearest neighbors. }
		\label{fig:various_outs}
	\end{figure*}
	
	\begin{figure*} [ht]
		\centering
		\includegraphics[width=0.99\linewidth]{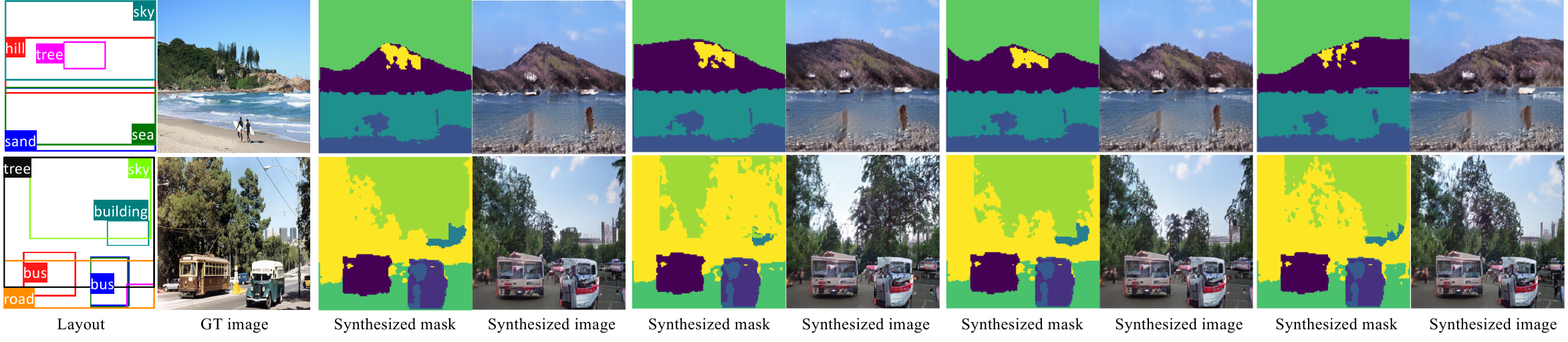}
		\caption{\textbf{Fine-grained object-level style control} w.r.t. a specific object instance in our LostGAN-V2 $256\times 256$: The \textit{hill} bounding box and the \textit{tree} bounding box are selected as the object instance to vary in the two rows respectively. We can observe sufficient changes of the two selected object instances, while the style of remaining objects are retained. Another example of person is shown in Fig.~\ref{fig:teaser}.}
		\label{fig:generated_masks}
	\end{figure*}
	
	\textbf{Style controllability} of our model is shown in Fig.~\ref{fig:various_outs} by synthesizing images with different visual appearance for a given layout, encoded by different $(z_{img}, Z_{obj})$ samples, while preserving objects at desired locations. The AlexNet-pool5 nearest neigbhors in the training dataset show that the synthesized images are not observed to suffer from overfitting.

	\begin{figure}[!h]
		\centering
		\includegraphics[width=0.99\linewidth]{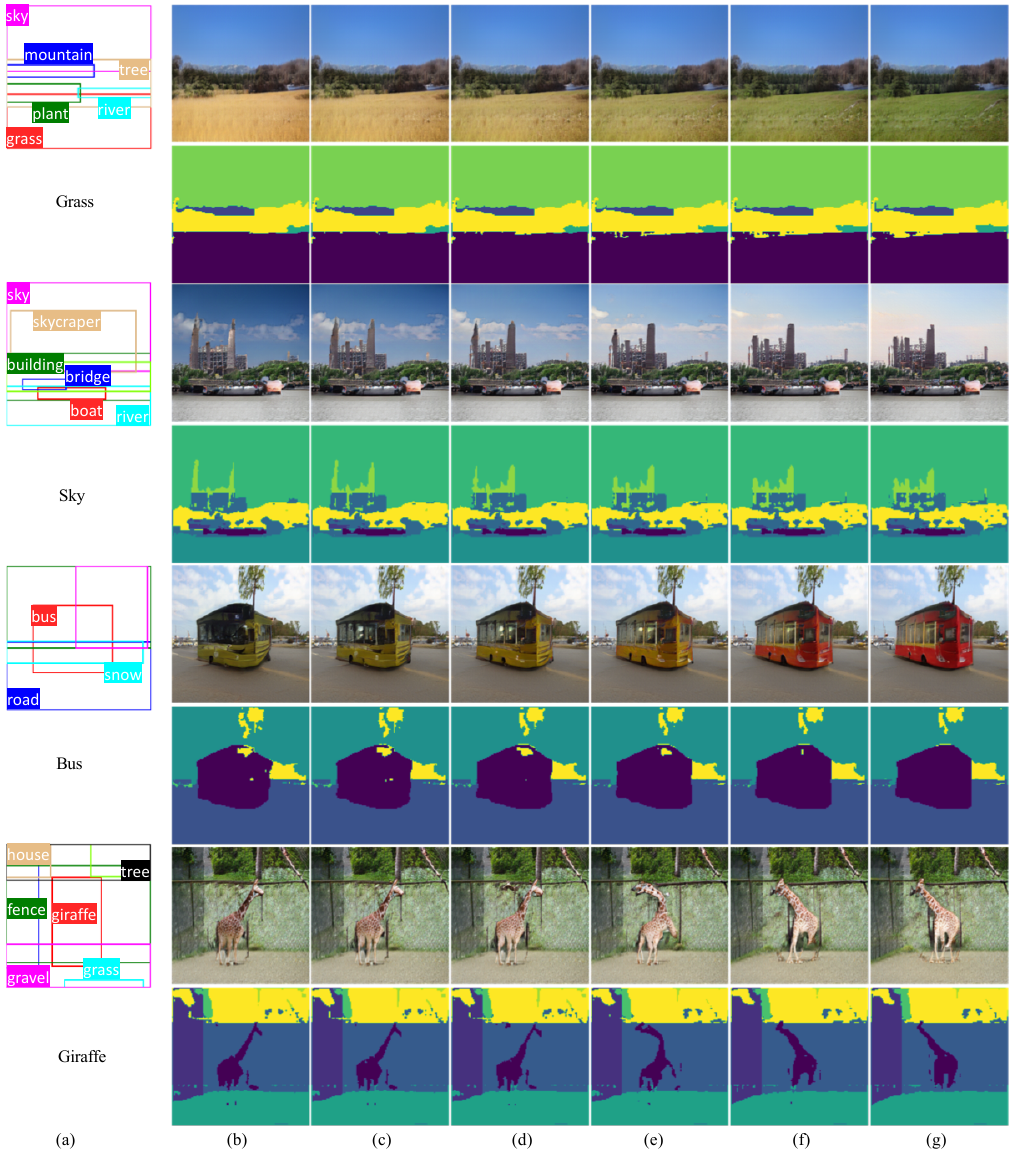}
		\caption{\textbf{Fine-grained object-level style control} in our LostGAN-V2: We use linear interpolation of object instance style codes, $z_{obj_i}$. The objects-of-interest are grass, sky, bus and giraffe respectively. For each layout, we first generate two images in (b) and (g) using two different $z_{obj_i}$ samples, $z_{obj_i}^1$ and $z_{obj_i}^2$,  while $(z_{img}, Z_{obj}\setminus \{z_{obj_i}\})$ are kept the same. Then, we synthesize 4 images in from (c) to (f) using object style codes linearly interpolated from $z_{obj_i}^1$ and $z_{obj_i}^2$. See text for details.}
		\label{fig:instance_style}
	\end{figure}
	
	\textbf{Fine-grained object-level style controllability} of our model is further  shown in Fig.~\ref{fig:generated_masks} and Fig.~\ref{fig:instance_style}. Fig.~\ref{fig:generated_masks} shows the controllability and the resulting diversity by changing the style latent code of a specific object instance.  Fig.~\ref{fig:instance_style} shows the effects of gradually morphing styles of one instance in different synthesized images. 
	
	We have three observations as follows: (i) Our LostGAN-V2 is capable of disentangling the styles in synthesis at the object instance level with sufficient diversity induced for a specific object instance. This is controlled by the object instance specific layout-aware learning of the affine transformation parameters (Section~\ref{sec:isla-norm}).  (ii) Our LostGAN-V2 is capable of handling the style morphing at the object instance level. (iii) Our LostGAN-V2 is capable of inducing semantically meaningful interpretations for the latent style codes via the proposed ISLA-Norm. For the ``stuff" such as grass and sky in the left of Fig.~\ref{fig:instance_style}, the change of an object style code does not affect its own object mask and the styles of remaining objects. For the ``things" such as bus and giraffe in the right of Fig.~\ref{fig:instance_style}, our LostGAN-V2 shows some interesting results. When linearly interpolating the latent style codes, for the bus example, the generator mainly changes the appearance according the change of latent codes in the morphing, while for the giraffe example, the generator changes both the appearance (slightly) and the pose. The learned object masks support these. So, it seems that the generator learns to understand the mixed semantic meanings of object latent style codes.

	\subsubsection{Results of the layout-to-mask module in LostGANs}\label{sec:layout2mask}
	Fig.~\ref{fig:generated_masks} shows examples of learned masks, in which even for complex scene with multiple overlapping objects, synthesized images and learned masks are consistent and semantically reasonable. Compared to the input bounding boxes, the learned masks help reduce the semantic gap in layout-to-image. Those masks are learned jointly with image synthesis in a single generator in a weakly-supervised manner, verifying our proposed pipeline of simultaneously learning layout-to-mask-to-image.

	\begin{figure}%[!ht]
		\centering
		\includegraphics[width=\linewidth]{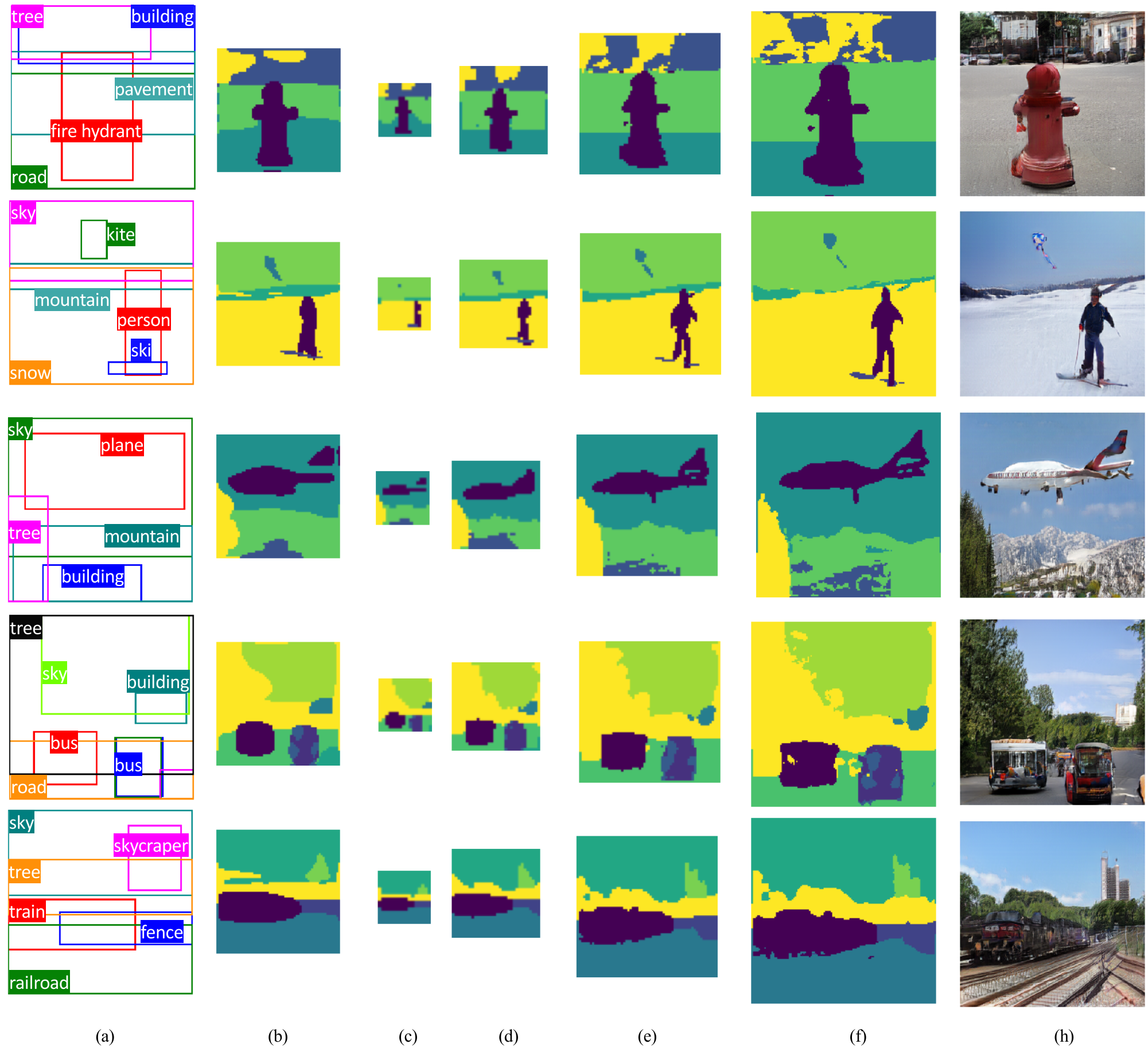}
		\caption{Examples of mask refinement in the generator. (a) Layouts, (b) Initial masks generated form the joint label and style encoding, (c-f) Mask refinement using masks generated from feature maps at different stages in the generator, (h) Synthesized images.}
		\label{fig:mask_refine}
	\end{figure}

	Fig.~\ref{fig:mask_refine} shows examples of mask refinement in the process of generation. The initial mask generation can produce reasonably good results, which are refined in the cascade of integrating masks learned from feature maps, especially for object boundaries (\eg, comparing (b) and (f)). This mask refinement is one of the main technical improvement between our LostGAN-V1 and LostGAN-V2, which also verifies the overall improvement by the LostGAN-V2 in experiments.

	\subsection{Results of layout semi-supervised LostGANs} \label{sec:semi}
	We test the semi-supervised LostGAN-V2 at the resolution of $128\times 128$ in the COCO-Stuff dataset~\cite{caesar2018coco}. We randomly and evenly split the training dataset into two subsets, denoted by $D_1$ and $D_2$ respectively. We discard the bounding box annotations in $D_2$. Using $D_1$, we train a Faster-RCNN~\cite{FasterRCNN} detector for the entire $171$ categories including both ``things" and ``stuff". Then, we run the trained Faster-RCNN detector in $D_2$ to generate detection results with a threshold $\tau=0.5$. The AP$^{bb}_{50}$ in $D_2$ is 68.9\%.
	
	Table~\ref{tab:semi} shows the results. We test four settings: (i) LostGAN-V2 trained with $D_1$; 
	% (ii) LostGAN-V2 trained with $D_1$ for both the generator and discriminator and with $D_2$ for the discriminator; 
	(ii) LostGAN-V2 trained with $D_1$ and $D_2$ together with the Faster-RCNN detection results using the threshold $0.5$; (iii) LostGAN-V2 trained with $D_1$ and $D_2$ together with the Faster-RCNN detection results using the threshold $0.5$ and Eqn.~\ref{eq:lossD1}. 
	
	\begin{table}[!h]
		\centering
		\caption{Comparisons of semi-supervised LostGAN-V2 128$\times$ 128. See text for details.}
		\label{tab:semi}
		\begin{tabular}{ p{3.5cm}|c|c|c } 
			\hline 
			\textbf{Setting} & IS$\uparrow$ & FID$\downarrow$ & DS$\uparrow$ \\ \hline 
			(i) $D_1$ & 12.79$\pm$0.27 & 32.77 & 0.46 $\pm$ 0.09 \\ \hline
			%  (ii) $D_1$+$D_2:$ $\mathcal{D}$ & 13.2 & 27.2 & 0.45 $\pm$ 0.9 \\ \hline 
			(ii) $D_1$+$D_2:$ Detection($0.5$) & 13.41$\pm$0.48 & 28.20 & 0.46 $\pm$ 0.09 \\ \hline 
			(iii) $D_1$+$D_2:$ Detection($0.5$) + Eqn.~\ref{eq:lossD1} & \textbf{13.90$\pm$0.38} & \textbf{25.87} & 0.46 $\pm$ 0.09 \\ \hline 
			Fully-supervised & 14.21$\pm$0.40 & 24.76 & 0.45 $\pm$ 0.09 \\ \hline 
		\end{tabular}
	\end{table}

	First, the diversity score (DS) is not affected since it reflects the variations between different synthesized images and the style control of our LostGAN is not directly associated with the number of data (Fig.~\ref{fig:norm} and Eqn.~\ref{eq:jointstyle}). Then, in terms of IS and FID, LostGAN-V2 trained with half data (i) has worse performance, but not very significantly. LostGAN-V2 under the setting (iii) obtains results comparable to the fully-supervised counterpart. This may indicate \textbf{the bottleneck of layout-to-image synthesis} is not the amount of annotated bounding boxes in training, but the capabilities of inferring better object masks on-the-fly. This is consistent with observations from existing work:  BigGANs~\cite{brock2018large} and  StyleGANs~\cite{karras2018style} %~\footnote{We view the StyleGAN as an implicitly conditional image synthesis framework since only one category is usually handled in training} 
	have shown great results of image synthesis using simpler conditions than layouts. GauGANs~\cite{park2019semantic} also have shown great results of image synthesis conditioned on annotated semantic maps.  
	
	\textit{Remarks:} We train the Faster-RCNN and the LostGAN individually for simplicity. It seems promising that we can train them jointly. In the meanwhile, we also can explore how to leverage the LostGAN to help the Faster-RCNN by generating more data similar in spirit to the CAS evaluation in Table~\ref{tab:classification_score}. Thus, it is possible to form a three-player minmax game such that both the Faster-RCNN and the LostGAN can benefit each other under a semi-supervised learning settings. We leave this for future work.

	\subsection{Comparisons with the GauGAN}\label{sec:compare_spade}
	% The GauGAN~\cite{park2019semantic} is one of the state-of-the-art semantic-map-to-image synthesis methods. The GauGAN and our LostGAN-V1~\cite{LostGAN} are concurrent works on developing spatially-adaptive feature normalization schema for condition controllability in conditional GANs. The GauGAN utilizes annotated semantic label maps in training and inference, while our LostGANs learn a mapping for layout-to-mask interacting with the generator network. 

	We conduct experiments for the methods presented in Section~\ref{sec:compare_spade}.
	
\begin{figure}[!h]
		\centering
		\includegraphics[width=0.99\linewidth]{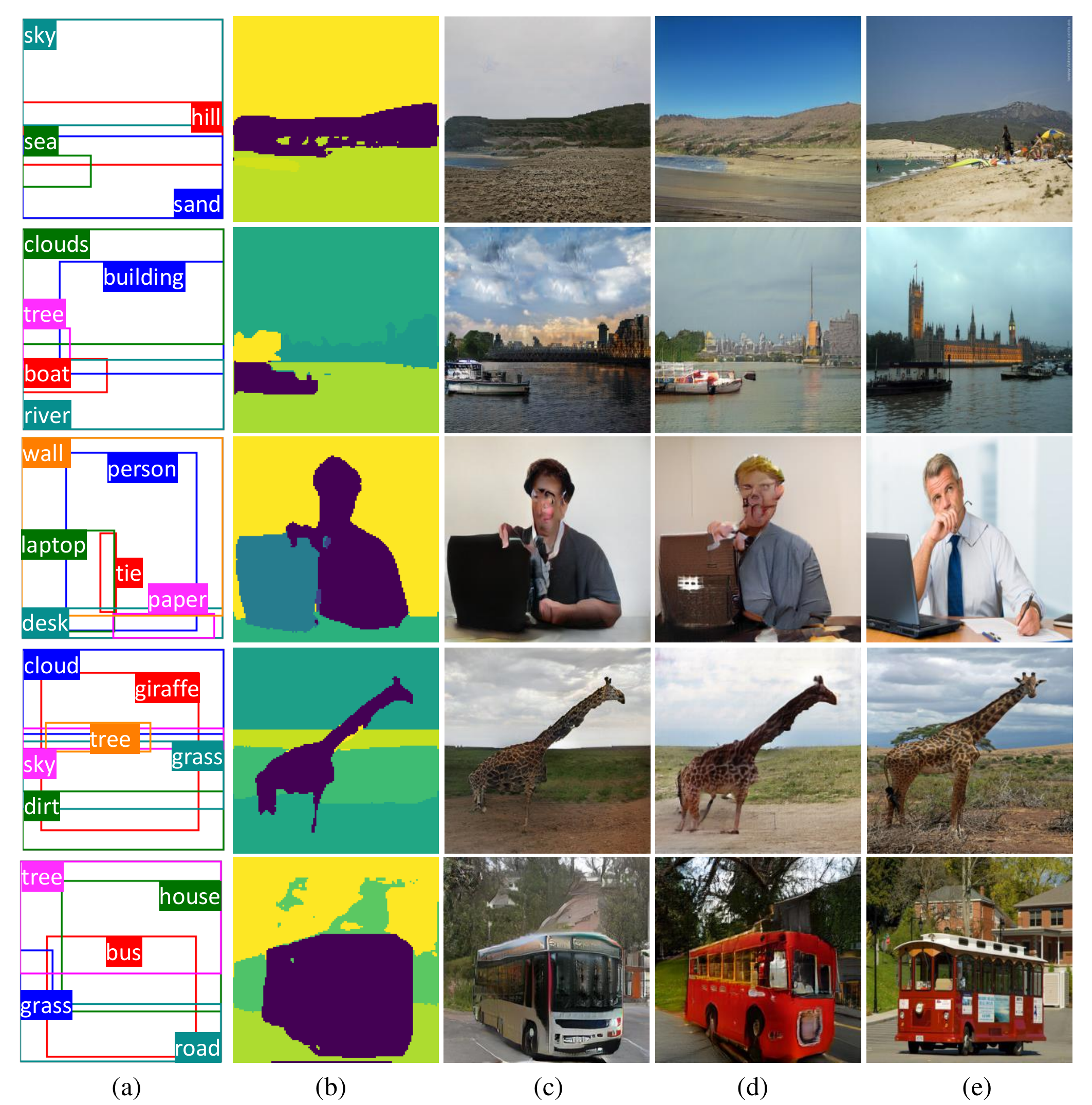}
		\caption{Comparison of our method and GauGAN~\cite{park2019semantic} at the resolution of $256\times 256$. (a) Input Layouts, (b) Masks learned by our model, (c) Synthesized images by GauGAN using the masks in (b), (d) Generated images by our LostGAN-V2, (e) Ground Truth images. Our model achieves comparable visual performance with GauGAN, which is trained with supervision of ground truth masks. See text for details.}
		\label{fig:compare_spade}
	\end{figure}

	Table~\ref{tab:spade} shows the comparisons in terms of IS, FID and DS. First, with the post-hoc integration, the GauGAN obtains slightly better IS and FID than our LostGAN-V2, while our LostGAN-V2 achieves better DS. Fig.~\ref{fig:compare_spade} shows some examples, from which we can see the generator in our LostGAN-V2 works reasonably good, comparing to the GauGAN that are trained with ground-truth masks. Second, for the end-to-end integration between the layout-to-mask component and the mask refinement component in our LostGAN-V2 and the SPADE in the GauGAN, our LostGAN-V2 obtains slightly better performance with one possible explanation stated in Section~\ref{sec:compare_spade}. Third, the vanilla GauGAN that uses ground-truth semantic masks in both training and testing obtains significantly better FID. 
	The results further verify the importance of object masks in learning layout-to-image generator models, and show the effectiveness of the proposed layout-to-mask module in our LostGANs. Along with our layout semi-supervised LostGAN results, developing better layout-to-mask modules will be one of the main directions to be addressed in future work. 
	
	\begin{table}[t]
		\centering
		\caption{Quantitative comparisons using the Inception Score (IS, higher is better), the FID (lower is better) and Diversity Score (DS, higher is better) evaluation on COCO-Stuff dataset at the resolution of $256\times 256$. See text for details.}
		\label{tab:spade}
		\resizebox{0.99\linewidth}{!}{
			\begin{tabular}{p{4cm}|c|c|c}
				\hline
				\textbf{Methods} & \textbf{IS}$\uparrow$ & \textbf{FID}$\downarrow$ & \textbf{DS}$\uparrow$ \\ \hline
				GauGAN (w/ GT masks)~\cite{park2019semantic} & -  & {22.6}  & -  \\ \hline\hline 
				GauGAN~\cite{park2019semantic} + Our Inferred Masks & {19.35 $\pm$ 0.73}  & 41.11  & 0.38 $\pm$ 0.12  \\ \hline\hline 
				Our LostGAN-V2 & \textbf{18.01$\pm$0.50} & \textbf{42.55} & \textbf{0.55$\pm$0.09} \\
				\hline 
				LostGAN-V2 + SPADE~\cite{park2019semantic} (end-to-end) & 16.37 $\pm$ 0.34 & 46.08& 0.50 $\pm$ 0.11 \\ \hline
		\end{tabular}}
	\end{table}
	
	\subsection{Ablation study}
	\subsubsection{Effects of different components in LostGANs}
	We test the effects of four different components in our LostGAN-V2. Two components in the generator network $\mathcal{G}$: the layout-to-mask component and the mask refinement component. Two components in the discriminator network $\mathcal{D}$: the image head classifier and the object head classifier. Due to the computational budget requirement, we do not perform combinatorial ablation studies between the four components. 
	
	\begin{table}[!ht]
		\centering
		\caption{Effects of four components in LostGAN-V2 128$\times$ 128. See text for details.}
		\label{tab:abla_study}
		\begin{tabular}{ c|c|c|c } 
			\hline 
			\textbf{Components} & \textbf{IS$\uparrow$} & \textbf{FID$\downarrow$}  & \textbf{DS$\uparrow$}\\ \hline 
			w/o layout-to-mask in $\mathcal{G}$ & 12.91$\pm$0.23 & 28.51 & 0.45$\pm$0.09 \\ \hline 
			w/o mask refinement in $\mathcal{G}$  &13.75$\pm$0.49 &26.11 &0.46$\pm$0.09 \\ \hline
			w/o image head in $\mathcal{D}$ & 13.85$\pm$0.31 & 25.96 &0.43$\pm$0.10 \\\hline 
			w/o object head in $\mathcal{D}$ & 9.51$\pm$0.22 & 57.06 &0.55$\pm$0.11 \\\hline
			Full LostGAN-V2 & 14.21$\pm$0.40 & 24.76 & 0.45$\pm$0.09 \\ 
			\hline 
		\end{tabular}
	\end{table}
	
	Table~\ref{tab:abla_study} shows the comparisons between the four components individually. The comparisons are done at the resolution of 128$\times $128. Overall, each of the four components has a significant effect on the synthesis results, which supports the proposed design of our LostGANs-V2. The component that affects the results most is the object head classifier in the discriminator network. This is aligned with the GAN setting: the discriminator network provides learning signals to the generator network, and for layout-to-image synthesis the object level learning signals are of the most importance. 
	
	\begin{table}[!ht]
		\caption{Effects of the mask refinement in our LostGAN-V2 $256\times256$ in COCO-Stuff. $m_0$ represents initial masks generated from the joint label and style encoding. $m_i$ represents the refined masks at the $i$-th stage of the generator. See text for details.}
		\label{tab:mask_study}
		\centering
		\begin{tabular}{c|cc}
			\hline 
			Mask branch & IS$\uparrow$ & FID$\downarrow$ \\ \hline
			$m_0$  & 16.68 $\pm$ 0.42 & 48.54 \\
			$m_0$ + $m_1$ & 14.14 $\pm$ 0.33 & 63.96 \\
			$m_0$ ... $m_2$  & 17.10 $\pm$ 0.56 & 48.94 \\
			$m_0$ ... $m_3$ & 17.46 $\pm$ 0.34 & 44.38 \\
			$m_0$ ... $m_4$  & 17.51 $\pm$ 0.41 & 42.49 \\
			$m_0$ ... $m_5$  & 18.01 $\pm$ 0.50 & 42.55 \\
			\hline
		\end{tabular}
	\end{table}
	
	\subsubsection{The iterative mask refinement component}
	We conduct an ablation study on the iterative mask generation component (Fig.~\ref{fig:mask_refine}) to investigate their effects. After training, we compare the performance of different models with some of mask refinement stages removed. As shown in Table~\ref{tab:mask_study}, the last row shows the full model with all the mask components, $m_0\cdots m_5$. In a backward way, if we remove the mask refinement stage by stage in the generator, the performance (Inception Score and FID) are indeed negatively affected. However, if we remove all the mask refinement stages and only use the initial masks, the performance is better than the model with  mask refinement in the first stage, $m_0+m_1$. One potential reason is that the resolution of first stage is very low, from which the learned masks may overlook objects of small sizes and introduce artifacts in the predicted masks. After observing this in the ablation study, we re-trained a model without using $m_1$ in COCO-Stuff and did not observe performance improvement, so we did not re-train all the models used in our experiments.

	\section{Conclusions and Discussions}\label{sec:conclusion}
	This paper studies the generative learning problem of layout-to-image with a focus on controllable image synthesis from reconfigurable structured layouts and styles. This paper first presents an intuitive pipeline of learning layout-to-mask-to-image. Then, it presents a layout- and style-based architecture  for generative adversarial networks (termed LostGANs). The proposed LostGAN can be trained end-to-end to generate images from reconfigurable layout and style with strong style and layout controllability at both image and object levels. Our proposed LostGAN also can  learn fine-grained object masks in a weakly-supervised manner to bridge the gap between layouts and images by a novel object instance-sensitive layout-aware feature normalization (ISLA-Norm) scheme. State-of-the-art performance is obtained in the COCO-Stuff and Visual Genome datasets. 
	
\begin{figure}[!ht]
		\centering
		\includegraphics[width=0.95\linewidth]{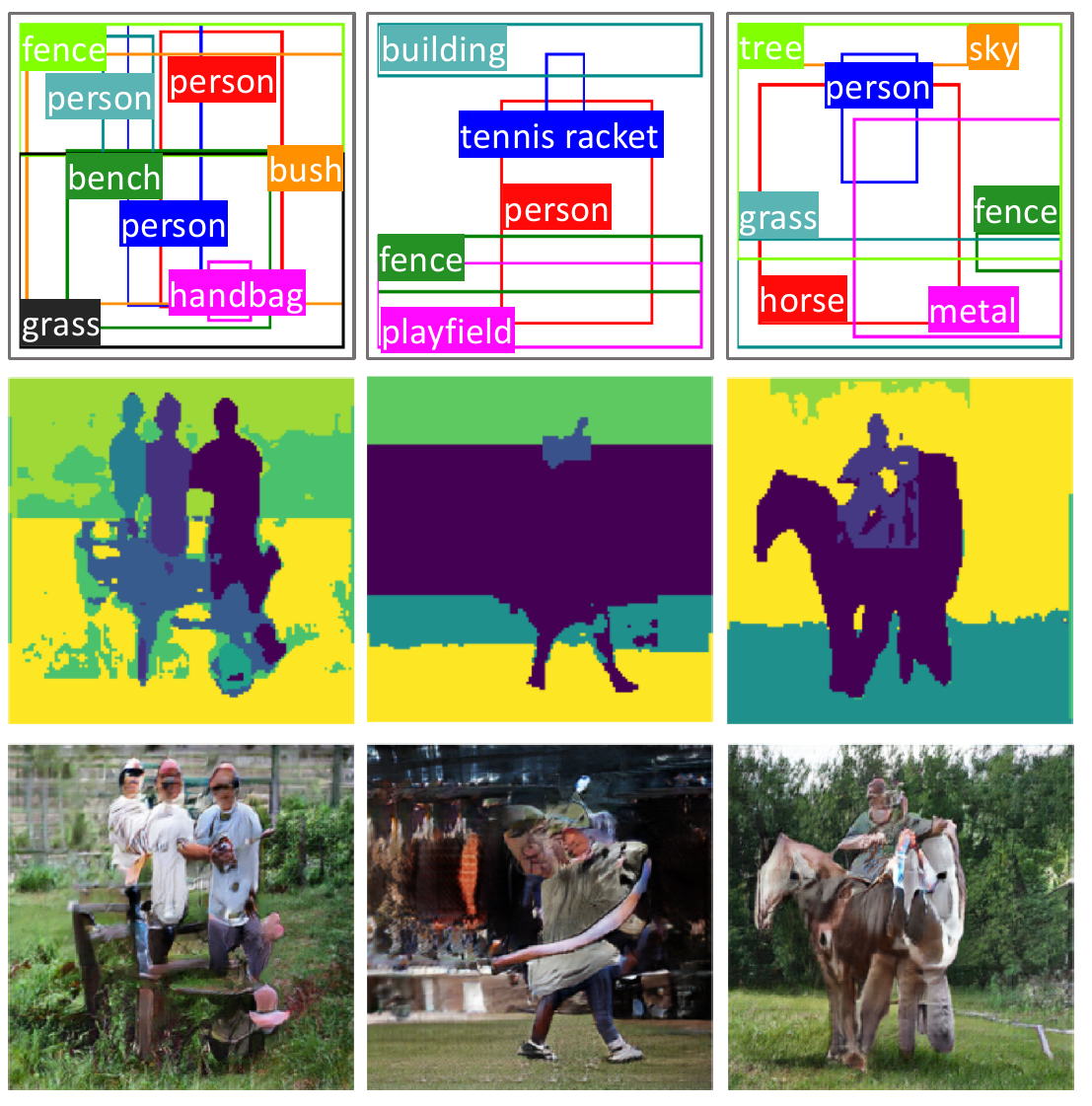}
		\caption{Examples of failure cases of our LostGAN-V2 $256\times 256$.}
		\label{fig:failure_case}
	\end{figure}

	\textbf{Discussions.} The generative learning problem of layout-to-image synthesis is still at a early stage of development in terms of synthesizing high-fidelity images, compared to the results of BigGANs~\cite{brock2018large} in ImageNet and StyleGANs~\cite{karras2018style} for faces. Overall, we can observe the quality of image generation from layout is still not sufficiently good, especially for articulated objects (such as people) and fine-grained object-object interactions at high resolution (\eg, examples in Fig.~\ref{fig:failure_case}). In Fig.~\ref{fig:failure_case}, we observe that our proposed model is not capable of capturing interactions between person and small objects, \eg, person and tennis racket in the middle column. From the learned label maps, we also can see why the model can not synthesize visually good images. We leave this to our future work by investigating methods of learning fine-grained part-level masks. 
	
	In the meanwhile, we also note that the differences between the goals of BigGANs and StyleGANs and those of controllable layout-to-image synthesis are non-trivial. For example, we can use a trained BigGAN to generate cat images, and as long as the generated images look realistic and sharp with one or more than one cats, we shall think it does a great job (without requiring how many cats should appear and where they should be). Similarly, we can train a StyleGAN to generate face images, and we shall be happy if realistic and sharp face images are generated with a natural style (\eg, smiling or sad). Controllable layout-to-image synthesis has more fine-grained requirements, which is relatively more challenging similar in spirit to the classic constraint-satisfaction problems in AI. 
	Those being said, based on the promising results of GauGANs~\cite{park2019semantic} using annotated semantic maps in image synthesis, we think the proposed layout-to-mask-to-image pipeline and LostGANs worth further explorations of seeking more powerful weakly-supervised learning of layout-to-mask. For example, we can develop more sophisticated mask generator networks and the ``ToMask" modules in Fig.~\ref{fig:norm}. We also should explore different consistency constraints between the ``ToMask" modules along the layers of the generator network, similar to the recently proposed PointRend method for improving Mask-RCNNs~\cite{PointRend}. 
	In addition to improve the layout-to-mask generation, it is an important direction of designing better discriminator networks that provide better fine-grained supervision signals for the generator network. We leave those for the future work.
	
	% Along with developing more powerful layout-to-image synthesis systems, it entails more work on developing better evaluation metrics for layout-to-image synthesis.  
	
	\section*{Acknowledgements}
	This work was supported in part by NSF IIS-1909644 and ARO Grant W911NF1810295. The views presented in this paper are those of the authors and should not be interpreted as representing any funding agencies.
	%NSF IIS-1822477, NSF CMMI-2024688, NSF IUSE-2013451 and DHHS-ACL Grant 90IFDV0017-01-00
	\bibliographystyle{IEEEtran}
	\bibliography{ref}
	
	\vskip -1.6\baselineskip plus -1fil
	
	% 	\subfile{sections/biographies}
	% \subfile{bio}
	% \vspace{-8mm}
	% \newpage
	\begin{IEEEbiography}[{\includegraphics[width=1in,height=1.25in,clip,keepaspectratio]{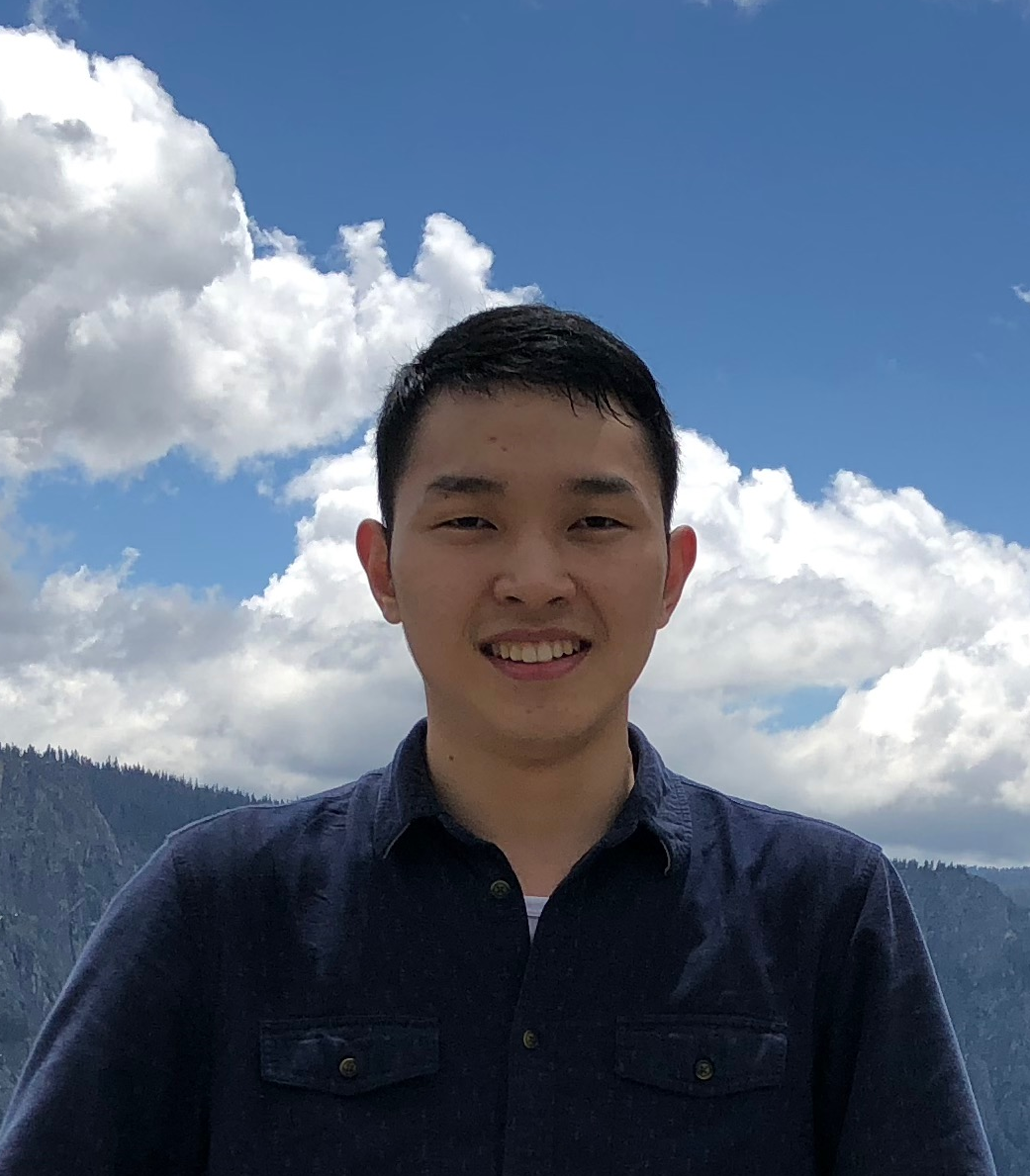}}]{Wei Sun}
		was recently graduated with a Ph.D. degree in the Department of Electrical and Computer Engineering at NC State University (NCSU). He is now a research scientist at Facebook. 
		He received the B.S. degree in Department of Physics from Nanjing University, Nanjing, China in 2015. His research interests include deep generative learning and its integration in image parsing.
	\end{IEEEbiography}

	\vskip -1.6\baselineskip plus -1fil
	
	\begin{IEEEbiography}[{\includegraphics[width=1in,height=1.25in,clip,keepaspectratio]{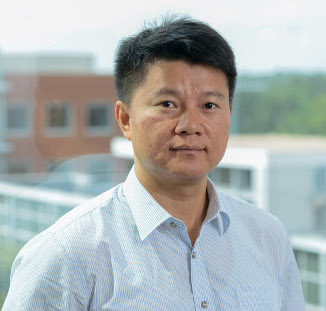}}]{Tianfu Wu}
		is an assistant professor in the Department of Electrical and Computer Engineering at NC state university (NCSU). He received his Ph.D. in statistics from UCLA under the supervision by Prof. Song-Chun Zhu.	His research focuses on computer vision, often motivated by the task of building explainable and improvable visual Turing test and robot autonomy through life-long communicative learning. To accomplish his research goals, he is interested in pursuing a unified framework for machines to ALTER (Ask, Learn, Test, Explain and Refine) recursively in a principled way.
	\end{IEEEbiography}

\end{document}